\documentclass[10pt,twocolumn,letterpaper]{article}

\usepackage{iccv}
\usepackage{times}
\usepackage{epsfig}

\usepackage{graphicx}
\usepackage{amsmath}
\usepackage{amssymb}
\usepackage{booktabs}
\usepackage{multirow}
\usepackage{tabularx}
\usepackage[table]{xcolor}
\usepackage{enumitem}
\usepackage[format=plain,labelformat=simple,labelsep=period,font=small,compatibility=false]{caption}
\usepackage[markup=underlined]{changes}

\newcolumntype{C}{>{\centering\arraybackslash}X}

\usepackage[pagebackref=true,breaklinks=true,letterpaper=true,colorlinks,bookmarks=false]{hyperref}

\usepackage[capitalize]{cleveref}
\crefname{section}{Sec.}{Secs.}
\Crefname{section}{Section}{Sections}
\Crefname{table}{Table}{Tables}
\crefname{table}{Tab.}{Tabs.}

\definecolor{3}{RGB}{255, 255, 200}
\definecolor{2}{RGB}{255, 220, 200}
\definecolor{1}{RGB}{255, 181, 163}
\newcommand{\cc}[1]{\cellcolor{#1}}

\newcommand{\gitlink}{\href{ https://yiyuzhuang.github.io/NeAI/}{ https://yiyuzhuang.github.io/NeAI/}}

\iccvfinalcopy 

\def\iccvPaperID{****} 

\begin{document}


\title{A Pre-convoluted Representation for \\ Plug-and-Play Neural Ambient Illumination}


\author{\normalsize 
Yiyu Zhuang$^{1}$\footnotemark[1]\ , Qi Zhang$^{2}$\footnotemark[1]\ , Xuan Wang$^{2}$, Hao Zhu$^{1}$, Ying Feng$^{2}$, Xiaoyu Li$^{2}$, Ying Shan$^{2}$, Xun Cao$^{1}$ \\
\normalsize $^{1}$ Nanjing University, Nanjing, China \ \ \ \ 
\normalsize $^{2}$ Tencent AI Lab, Shenzhen, China\\
{\tt\small \gitlink }
}
%
%


\twocolumn[{%
\renewcommand\twocolumn[1][]{#1}%
\maketitle

\begin{center}
    \centering
    \captionsetup{type=figure}
 \includegraphics[width=0.98\hsize]{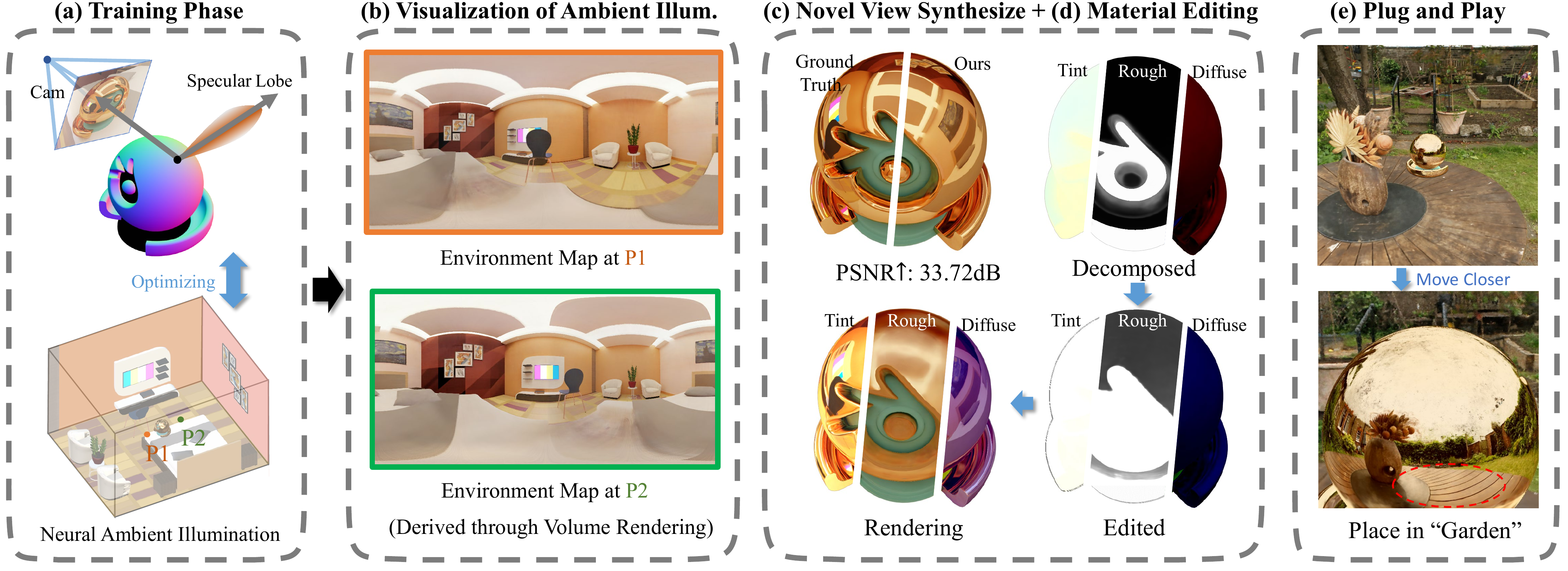}
    \caption{We propose a ``plug-and-play'' neural ambient illumination (NeAI) that uses the volumetric radiance fields to express the sample of the surrounding 3D environments as a light emitter and re-renders a new object under arbitrary NeRF-style environments naturally. (a) Using a set of images and masks, our method first optimizes the geometry and then jointly optimizes the NeAI and an object's materials in two stages.
    (b) Illustrates the environment map at arbitrary samples in 3D scenes via volume rendering, which addresses the environment occlusion and directional lighting naturally. The proposed method provides photo-realistic novel views (c) and visually reasonable material editing (d). More importantly, (e) applies our model to a pre-trained NeRF scene, and produces realistic specular reflections with nice directional illumination. Notably, the Fresnel effect of the table (\textcolor{red}{circle}) is exactly reproduced, which is almost impossible for the previous lighting representation. }
    \label{fig:teaser}
\end{center}%
}]

\renewcommand{\thefootnote}{\fnsymbol{footnote}} 
\footnotetext[1]{Authors contributed equally to this work.}

\begin{abstract}
    Recent advances in implicit neural representation have demonstrated the ability to recover detailed geometry and material from multi-view images. However, the use of simplified lighting models such as environment maps to represent non-distant illumination, or using a network to fit indirect light modeling without a solid basis, can lead to an undesirable decomposition between lighting and material. To address this, we propose a fully differentiable framework named neural ambient illumination (NeAI) that uses Neural Radiance Fields (NeRF) as a lighting model to handle complex lighting in a physically based way. Together with integral lobe encoding for roughness-adaptive specular lobe and leveraging the pre-convoluted background for accurate decomposition, the proposed method represents a significant step towards integrating physically based rendering into the NeRF representation. The experiments demonstrate the superior performance of novel-view rendering compared to previous works, and the capability to re-render objects under arbitrary NeRF-style environments opens up exciting possibilities for bridging the gap between virtual and real-world scenes. The project and supplementary materials are available at \textit{\textcolor{cyan}{\gitlink}}.

\end{abstract}

\section{Introduction}
\label{sec:intro}

Modeling and representing the ambient illumination from multi-view images is the fundamental issue that has been extensively studied throughout the development of rendering algorithms in computer vision and graphics\cite{park2020seeing, yao2022neilf, zhang2022modeling}. This task is inherently related to materials decomposition, since the observed scene appearance is affected by the interactions between ambient illumination and scene materials. It has drawn significant attention in this era of blowout VR and AR applications, where there is a high demand for photo-realistic rendering of the scene with visually natural illumination in a realistic environment. However, this problem is hard to solve because the ambient illumination is of high-dimensional information and strongly coupled with materials. 

Recent methods employ approximated illumination representations (\eg, environment map \cite{zhang2021nerfactor}, and spherical Gaussian (SG) models \cite{zhang2021physg, boss2021nerd, boss2021neural, zhang2022modeling}) to simplify the interaction between environment lighting and the object and reduce computational expense. Unfortunately, the assumption used in doing so is that the environmental lighting of the scene is infinitely far away. 
Neither the environment map nor SG models take the position of 3D environments into account so they are unable to handle environment occlusion and directional lighting practically. To address this problem, NeILF \cite{yao2022neilf} models an incident illumination map for each surface point to handle environment occlusion, indirect light and directional light. However, the constraint of spatial information in lighting is ignored, which is leading to worse material-lighting ambiguity under complex 3D environments. A common follow-up question to ask is: \emph{can we find an elegant representation to express the complex ambient illumination?} 

Recent advances in Neural Radiance Fields (NeRF) and its variants \cite{mildenhall2020nerf, barron2021mip, verbin2022ref, zhang2021nerfactor} have shown great potential to recover underlying scene proprieties (\eg geometry, materials, and lighting) from a set of images. NeRF uses a continuous volumetric function to represent the outgoing ray observed by a viewer.
The recent development of NeRF provides new possibilities to model complex ambient illumination. To the best of our knowledge, little in the literature has ever tapped into using NeRF to express lighting. 
By doing so we could achieve \emph{plug-and-play}: re-render an object with natural illumination of the NeRF-style real-world environment. However, directly gathering thousands of incoming rays through volume rendering to compute the color of each surface point is computationally expensive and may seem impossible.
In this paper, we present the \emph{Neural Ambient Illumination} (NeAI) to express the incoming ray of each surface point as the volumetric radiance fields (density and color) that have the efficient capability of handling environment occlusions and directional lighting naturally,  as shown in Fig.~\ref{fig:teaser}.
The first step of the proposed scheme is to acquire the object's 3D geometry. Fortunately, state-of-the-art geometry recovery methods\cite{wang2021neus, yariv2020multiview, yao2020blendedmvs, zhang2021physg} can provide almost accurate geometry information for most scenes. So in this work, we focus solely on optimizing ambient illumination and object material and further application. 
In particular, the pixel's specular color is equivalent to the interaction between object materials and the integral of incoming rays within the specular lobe, whose size is related to material roughness. Inspired by environment convolution maps used in traditional image-based lighting, we consider the rough refractive surfaces are inherently related to the convoluted background, although the background is ignored in most methods.
Our contributions are summarized as:
\begin{itemize}
	\setlength\itemsep{0pt}
    \item \emph{Neural Ambient Illumination} (NeAI) is proposed to expresse the radiance field of incoming rays such that treats each sample in the 3D scene as a light emitter.
    \item A fully \emph{differentiable rendering pipeline} is presented to seamlessly illuminate meshes using a NeRF-style environment.
    \item \emph{Integrated Lobe Encoding} (ILE) is proposed to featurize incoming rays within roughness-adaptive specular lobe to reduce computational cost.
    \item A \emph{multiscale pre-convoluted representation} for the background is proposed to assist in the decomposition of object materials and ambient illumination.
    
\end{itemize}
Extensive experiments and applications demonstrate the superior performance of our method. We strongly recommend browsing our project \textit{\textcolor{cyan}{\gitlink}} for better visualization.


\section{Related Work} \label{sec:related}
\noindent \textbf{Ambient Illumination Representation.} 
Ambient illumination representation is an essential component of photo-realistic rendering that aims to provide an immersive experience with visually natural illumination for a variety of view synthesis and relighting applications \cite{haber2009relighting, xu2018deep, gardner2017learning, bi2020deep, li2020inverse, boss2021nerd, srinivasan2021nerv, zhang2022modeling, yao2022neilf, hasselgren2022shape, boss2021neural, zhang2021nerfactor, zhang2021physg}.
Considering that the observed surface appearance in an image is the result of the interactions between ambient illumination and object materials, the ambient illumination is often jointly inferred with object materials from images, also known as inverse rendering \cite{sato1997object, marschner1998inverse, yu1999inverse, ramamoorthi2001signal, barron2014shape}. 
Since inverse rendering is well-known to be severely ill-posed, previous methods usually apply simplified material models \cite{zhang2021physg} and varying lighting conditions \cite{nam2018practical,bi2020deep3Dcapture, bi2020deep, li2022neural, yang2022s3nerf} to mitigate the problem. In this related work, we will specifically focus on techniques for ambient illumination representations.

The seminal work \cite{debevec1998rendering} proposes an omnidirectional radiance map, also known as \emph{environment map}, to represent ambient illumination, which can be applied to render novel objects into the scene realistically. Follow-up methods \cite{wen2003face, haber2009relighting, barron2014shape, valgaerts2012lightweight, song2019neural} use the environment map to handle inverse rendering problems naturally. Furthermore, given high-quality geometry, prior works \cite{lombardi2016radiometric, park2020seeing} factorize scene appearance into the diffuse image and the environment map from multi-view images. 
To be extended beyond a constant term, the environment map is expressed as \emph{spherical Gaussians} (SGs) formulation and integrated its product with surface material BRDF in the same representation to perform illumination calculations \cite{green2007efficient, wang2009all, zhang2021physg}. However, it has a significant approximation that light captured by the environment map is emitted infinitely far away. 


Considering the illumination from nearly all real-world light sources varies by direction as well as distance, \emph{global illumination representations} (\eg, ray-tracing \cite{miyazaki2007shape, srinivasan2021nerv} and path-tracing \cite{azinovic2019inverse, zhang2020path}) are proposed to use ray casting to express the interactions between ambient illumination and surface materials \cite{akenine2019real}. It is intrinsically described as a ray from a position to determine what objects are in a particular direction. However, global illumination representation is computationally expensive with the pre-computed process, and difficult to reconstruct to 3D real-world illumination for relighting without hand manner.
Our work takes inspiration from this line of work in graphics and presents a new ambient illumination representation. We represent the 3D sample of the surrounding environment as a light emitter, such that both position and direction of illumination are taken into account.

\noindent \textbf{Neural Radiance Fields.} Neural rendering \cite{mildenhall2020nerf, yariv2020multiview, liu2020neural, yariv2021volume, wang2021neus, oechsle2021unisurf, boss2021nerd, chen2022hallucinated, xie2023DINER, chen2023local}, the task of learning to recover the properties of 3D scenes from observed images, has seen significant success. In particular, Neural Radiance Fields (NeRF) \cite{mildenhall2020nerf} recover the radiance fields (volume density and view-dependent color) of a ray using a continuous volumetric function. Numerous works are extended from NeRF based on its continuous neural volumetric representation for generalizable models \cite{wang2021ibrnet, chen2021mvsnerf, johari2022geonerf, liu2022neural, huang2023LIRF}, dynamic scene \cite{du2021neural, li2022neural3dvideo, pumarola2021d}, non-rigidly deformable objects \cite{tretschk2021non, park2021nerfies, zhuang2022mofanerf}, camera model \cite{huang2022hdr, ma2022deblur, huang2023Inverting} and acceleration \cite{yu2021plenoctrees, reiser2021kilonerf, muller2022instant, yu2022plenoxels}. Recently, Mip-NeRF \cite{barron2021mip,barron2022mip} uses the integral along a cone instead of a ray to recover an anti-aliasing radiance field from a set of multi-scale downsampling images. Besides, Ref-NeRF \cite{verbin2022ref} is proposed for better reflected radiance interpolation. NeRF and its variants have demonstrated remarkable performance in rendering photo-realistic views, but they only model the outgoing radiance of the surface without considering the underlying interaction between ambient illumination and surface material.

Recent advances in differentiable rendering make it possible to reconstruct ambient illumination under casual lighting conditions. Specifically, PhySG \cite{zhang2021physg} and NeRD \cite{boss2021nerd} use SG representation to decompose the scene under complex and unknown illumination. Zhang \etal \cite{zhang2022modeling} model the indirect illumination via SGs without considering the environment occlusion. Neither the environment map nor SG models take the position of 3D environments into account so they are unable to handle environment occlusion and indirect lighting realistically. NeILF \cite{yao2022neilf} proposes the local environment map of each surface point for environment occlusion but ignores the distance of lighting. Overall, recent methods cannot construct a detailed illumination that takes into account both near-field lighting and environment occlusion. 
We propose neural ambient illumination representation that uses the volumetric radiance fields to express arbitrary illumination in the environment, such that the environment occlusion and directional lighting could be handled naturally.

\begin{figure*}
\centering
\includegraphics[width=\linewidth]{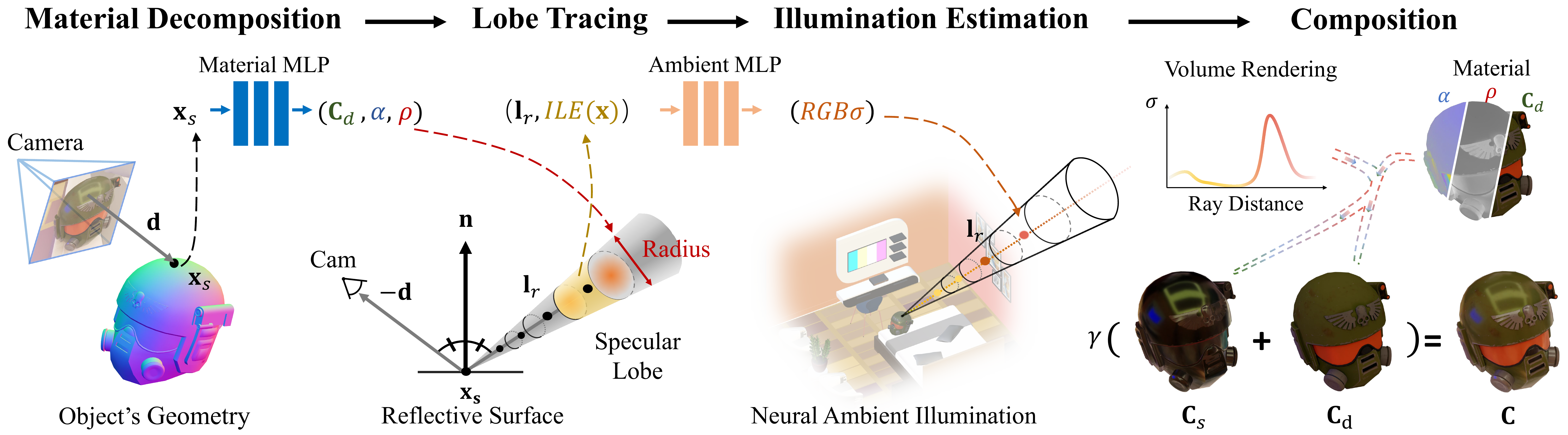}
\vspace{-0.2in}
\caption{The overview of the proposed method that decomposes materials $(\mathbf{C}_d, \alpha, \rho)$ of an object and NeAI radiance fields from a set of images and a reconstructed geometry. For a specific surface point $\mathbf{x}_s$, NeAI traces its incoming ray within a 3D specular lobe, whose center axis and width are defined by the reflection direction $\mathbf{l}_r$ and roughness $\rho$. We then feature samples along that lobe with our \emph{integral lobe encoding} (ILE) and feed into the ambient MLP to predict the color and volume density of each sample. Using volume rendering techniques, we integrate these values direction-wise multiplied by the material's function into the specular color $\mathbf{C}_s$. We combine this with the diffuse color $\mathbf{C}_d$ provided by the material MLP to render photo-realistic novel views. This rendering procedure is end-to-end differentiable, so we can jointly optimize our NeAI representation and object's materials.
 }
 \vspace{-0.1in}
\label{fig:pipeline}
\end{figure*}

\section{Method}
\label{sec:method}
Given a set of posed images of an object captured under static illumination, we learn to decompose the shape, material, and lighting. As previously mentioned, our main objective is  the representation of environment lighting and its interaction with the object surface. While the geometry reconstruction is an important aspect of the pipeline, it is \emph{not} the main focus of our work. Thus, we will introduce the core of our pipeline (see Fig. \ref{fig:pipeline}) after the object geometry has been reconstructed using external methods in stage one.



\subsection{Preliminaries}
Prior knowledge of NeRF and physically based rendering such as BRDF is recommended for readers to fully comprehend the modeling presented in this paper. \\
\noindent\textbf{NeRF.} NeRF \cite{mildenhall2020nerf} represents traditional discrete sampled geometry with a continuous volumetric radiance field (\ie density $\sigma$ and color $\mathbf{c}$).  Given a sampled point $\mathbf{x}$ along a single ray $\mathbf{r}$ originating at $\mathbf{o}$ with direction $\mathbf{d}$, a positional MLP predicts its corresponding density $\sigma(\mathbf{x})$, and a direction MLP outputs color $\mathbf{c}(\mathbf{x}, \mathbf{d})$ of that point along the ray direction. To render a pixel's color, NeRF casts a single ray $\mathbf{r}(t)=\mathbf{o}+t\mathbf{d}$ through that pixel and out into its volumetric representation, accumulates $(\sigma_i, \mathbf{c}_i)$ into a single color $\mathbf{C}(\mathbf{r})$ of the pixel via numerical quadrature \cite{max1995optical},
\begin{equation}\small
     \mathbf{C(r)}\!=\!\sum_{i} \exp\left({-\sum_{k=0}^{i-1}\sigma_{k}}\right)( 1-\exp(-\sigma_i) ) \mathbf{c}_i.
    \label{eq:vol_ren}
\end{equation}


\noindent\textbf{The rendering equation.} In contrast to NeRF, we replace the pixel's color of outgoing radiance from a surface point $\mathbf{x}_s$ along a view direction $\mathbf{d}$ with the diffuse color $\mathbf{C}_d$ of that point and an interaction (specular color $\mathbf{C}_s$) between the incoming radiance $\mathbf{L}_{in}$ of ambient illumination and scene material based on the rendering equation \cite{kajiya1986rendering},
\begin{equation}\small
    \mathbf{C}(\mathbf{x}_s, \mathbf{d}) = \mathbf{C}_d(\mathbf{x}_s) +  \int_{\Omega}{f(\mathbf{l}, -\mathbf{d},\mathbf{x}_s)\mathbf{L}_{in}(\mathbf{x}_{s},\mathbf{l})(\mathbf{l} \cdot \mathbf{n})}\mathrm{d}\mathbf{l},
    \label{eq:rendeing_int}
\end{equation}
where $\mathbf{n}$ and $\mathbf{l}$ denote the normal vector at $\mathbf{x}_s$ and the direction of $\mathbf{L}_{in}$ respectively. The $f$ is the bidirectional reflectance distribution function (BRDF) that focuses on the local reflectance phenomena. Eq. \eqref{eq:rendeing_int} integrate all incoming direction $\mathbf{l}$ on the hemisphere $\Omega$ where $\mathbf{l}\cdot \mathbf{n}>0$.

\subsection{Neural Ambient Illumination}
\label{sec:NeAI}
Although previous methods attempted to model diverse types of lighting in various ways \cite{zhang2021physg, zhang2021nerfactor, boss2021nerd, boss2021neural, zhang2022modeling, yao2022neilf}, there still remains a discrepancy between virtual and real-world scenes. However, NeRF has the potential to bridge this gap by enabling the accurate modeling of spatially and directionally varying light.

By expressing the \emph{neural ambient illumination} (NeAI) of an object directly as the continuous volumetric radiance field which includes the volume density and directional emitted radiance at any point in the 3D environment, we can achieve more precise modeling of the light.
Given a sample point $\mathbf{x}$ in the environment, we approximate this 5D volumetric radiance field function with the ambient MLP network $\mathcal{R}: (\mathbf{x},\mathbf{l}) \rightarrow (\mathbf{c}, \sigma)$,
where $\mathbf{l}$ is the direction of incoming ray pass through $\mathbf{x}$.
To further integrate the incoming radiance $\mathbf{L}_{in}$ of an incoming ray $\mathbf{r}(t)=\mathbf{x}_s+t\mathbf{l}$ to a surface point $\mathbf{x}_s$ along with the  direction $\mathbf{l}$, we accumulate the corresponding densities and directional emitted colors of $(\mathbf{r}(t), \mathbf{l})$ according to volume rendering,
\begin{equation}\small
\begin{aligned}
	\mathbf{L}_{in}(\mathbf{x_s}, \mathbf{l})=\int_{t_n}^{t_f} T(t) \sigma(\mathbf{r}(t)) \mathbf{c}(\mathbf{r}(t),\mathbf{l})\mathrm{d}t,&\\
	\textup{where} \: T(t)=\exp\left({-\int_{t_n}^t\sigma(\mathbf{r}(s))\mathrm{d}s}\right),&
\end{aligned}
\label{eq:L_i_int}
\end{equation}
where $T$ indicates the accumulated transmittance. In general, the near and far bounds $t_n$ and $t_f$ are ideally set to infinitely close to zero and infinitely distant, respectively. Eq. \eqref{eq:L_i_int} indicates that we treat each sample in the 3D environment as a light emitter.
This allows the proposed NeAI to model the directional emitted rays and environment occlusions of any static 3D environments. 

To define how light derived from a NeRF-style environment is reflected at an opaque surface, we parameterize the spatially-varying material of surface point $\mathbf{x}_s$ as roughness $\rho\in\left[0,1\right]$, diffuse color $\mathbf{C}_d\in\left[0,1\right]^3$ and specular tint $\alpha\in\left[0,1\right]$, which are output by a material MLP network (using a softplus activation), \ie, $\mathcal{M}: \mathbf{x}_s\rightarrow(\rho, \mathbf{C}_d, \alpha)$.
We assume that the BRDF $f$ is rotationally symmetric with respect to the reflection direction $\mathbf{l}_r$ around the specular lobe, such as Phong \cite{akenine2019real}. $\mathbf{l}_r$ is computed by $2(-\mathbf{d}\cdot\mathbf{n})\mathbf{n}+\mathbf{d}$.
Consequently, we approximate the BRDF as von Mises-Fisher (vMF) distribution which is defined on the unit lope with a normalized spherical Gaussian \cite{akenine2019real}, 
\begin{equation}\small
	f(\mathbf{l}, -\mathbf{d},\mathbf{x}_s)\!\approx\!G(\mathbf{l}, \mathbf{l}_r,\mathbf{x}_s) \!=\! \alpha\exp\left(\rho(\mathbf{x}_s)\left(\mathbf{l}\cdot\mathbf{l}_r-1\right) \right),
	\label{eq:ndf}
\end{equation}
where $\mathbf{l}$ and $\mathbf{l}_r$ are the unit vector, and the value is positively correlated to $\mathbf{l\cdot l_r}$. 
Specifically, $\mathbf{l}_r$ refers to the center axis of the lobe, and spatially-varying roughness $\rho(\mathbf{x}_s)$ controls its angular width (also called the concentration parameter or spread). 
Noting that, $\alpha$ could be considered as the lobe amplitude, which is learned from the material MLP.

By substituting Eq. \eqref{eq:ndf} and \eqref{eq:L_i_int} into  Eq. \eqref{eq:rendeing_int}, we obtain the specular term of Eq. \eqref{eq:rendeing_int} as:
\begin{equation}\small
	\mathbf{C}_s(\mathbf{x}_s, \mathbf{d})\!=\!\!\iint{
     \alpha e^{\rho\left(\mathbf{l}\cdot\mathbf{l}_r-1\right)}
	T(\mathbf{x}) \sigma(\mathbf{x}) \mathbf{c}(\mathbf{x},\mathbf{l})(\mathbf{l} \cdot \mathbf{n})\mathrm{d}\mathbf{x}\mathrm{d}\mathbf{l}}.
	\label{eq:our_render}
\end{equation}
According to  Eq. \eqref{eq:ndf},  a larger $\rho$ value corresponds to a rougher surface with a wider vMF distribution. Therefore, 
Eq. \eqref{eq:our_render} is equivalent to the integration of the radiance field of each sampled point in the specular lobe defined by the reflection direction. By doing so, we can more effectively and stereographically represent the reflection.

\subsection{Integrated Lobe Encoding}
\label{sec:ILE}
To better learn the high-frequency variation of NeAI related to roughness, we introduce a featurized representation, which we call an \emph{Integrated Lobe Encoding} (ILE), that efficiently and simply constructs positional encoding of all coordinates that lie within specular lobe around $\mathbf{l}_r$. Our ILE is inspired by IPE used in Mip-NeRF \cite{barron2021mip}, which enables the spatial MLP to represent volume density inside the cone along with view direction. 
We feature all coordinates inside a roughness-adaptive lobe which considers both the lobe width decided by the roughness and the vMF distribution correlated to the incoming direction.

We could divide the specular lobe of Eq. \eqref{eq:our_render} into a series of conical frustums. Similarly, we approximate this featurized procedure with a set of sinusoids via a multivariate Gaussian. Specifically, we compute the mean and covariance $(\mathbf{\mu}, \mathbf{\Sigma})$ of the conical frustum, which is obtained by $\mathbf{x}_s$ and $\mathbf{l}_r$. Note that the radius variance's part in $\mathbf{\Sigma}(\rho)$ is decided by the material roughness of the surface point, which is different with IPE \cite{barron2021mip}. Given that the integral value of Eq. \eqref{eq:our_render} is under a vMF distribution, it attenuate with the weights of $\mathbf{l}\!\cdot\!\mathbf{l}_r$.
Our $\mathrm{ILE}$ then formulates the encoding of those coordinates $(\mathbf{\mu}, \mathbf{\Sigma})$ within the conical frustum, 
\begin{equation}\small
        \mathrm{ILE} \!=\! \left\lbrace \!
        \begin{bmatrix}
            {\sin( \mathbf{\mu} )\exp(-2^{\ell-1})\mathrm{diag}(\mathbf{\Sigma}(\rho)))} \\
            {\cos ( \mathbf{\mu})\exp(-2^{\ell-1})\mathrm{diag}(\mathbf{\Sigma}(\rho)))}   
        \end{bmatrix}\!
        \right\rbrace_{\ell=0}^{L-1},
\end{equation}
These features encoded by $\mathrm{ILE}$ are used as input to the MLP network $\mathcal{R}$ to output the density and color of our NeAI. This encoding allows the MLP to parameterize the incoming illumination inside the roughness-varying specular lobe, whose strength is variable with the incoming direction under vMF distribution, to behave as an interpolation function. As a result, our ILE efficiently maps continuous input coordinates into a high-frequency space. Please refer to our \emph{supplement} for detail derivation. 

According to Eq. \eqref{eq:rendeing_int}, the pixel's color $\mathbf{C}$ captured by a camera is equivalent to the diffuse color $\mathbf{C}_d$ of that point and the volumetric integration $\mathbf{C}_s$ of the incoming rays within the specular lobe,
\begin{equation}\small \label{eq:object_render}
    \mathbf{C}(\mathbf{x}_s,\mathbf{d}) =
    \gamma\left(\mathbf{C}_d(\mathbf{x}_s) + \mathbf{C}_s(\mathbf{x}_s,\mathbf{d})\right),
\end{equation}
where $\gamma$ is a learned HDR-to-LDR mapping. We approximate it as the gamma correction with a learned parameter considering the transformation (\eg, exposure and white balance) learned into the radiance of incoming rays. 


\subsection{Multiscale Pre-convoluted Representation}
\label{sec:background}
Due the complexity of high-dimension representation of NeAI, it’s easy to integrate the diffuse color into ambient illumination, causing ambiguous decomposition. Consequently, we propose a multiscale pre-convoluted technique to introduce the background of the object to stabilize the convergence.
As shown in Eq. \eqref{eq:our_render}, the integral’s results of incoming rays within a specular lobe can be considered as convolution, and the width of the lobe is related to the roughness.  As the roughness increases, the ambient illumination is convoluted with more scattered samples within a wider lobe, creating blurrier reflections. By applying a set of different discrete Gaussian blur kernels to the background of the object, we obtain the pre-convoluted results of the incoming rays that correspond to different levels of roughness.

\begin{figure}
    \centering
    \includegraphics[width=1.0\linewidth]{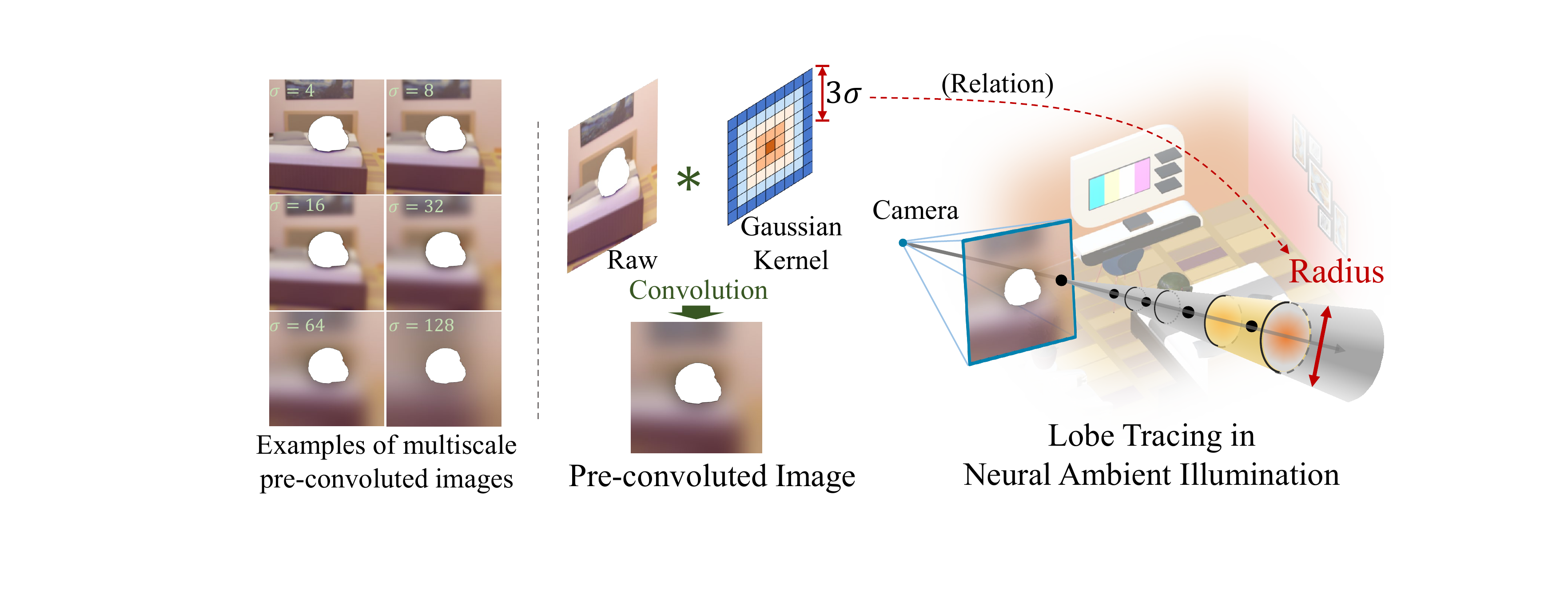}
    \caption{The pipeline demonstrates how pre-convolved images can be utilized to construct a pre-convoluted representation of illumination.}
    \vspace{-0.2in}
    \label{fig:pre-convoluted}
\end{figure}
In training phase as shown in Fig. \ref{fig:pre-convoluted}, the pixels of the background are evaluated through Eq. \eqref{eq:our_render} and used to supervise the decomposition. We manually set radius $r_p=3\sigma r_0$, $\alpha=1$, $\mathbf{C}_d=[0]^3$, where the $\sigma$ is the variance of the Gaussian kernel and  ${r_0}$ is the radius of the raw pixel size. 

This strategy looks similar to the pre-filtered environment map in CG rendering. However, we originally apply it to the neural rendering framework and take the view direction into account, which allows for more accurate and realistic specular reflections that are consistent with the object’s roughness level.




\begin{figure*}[t]
    \centering
    \includegraphics[width=\linewidth]{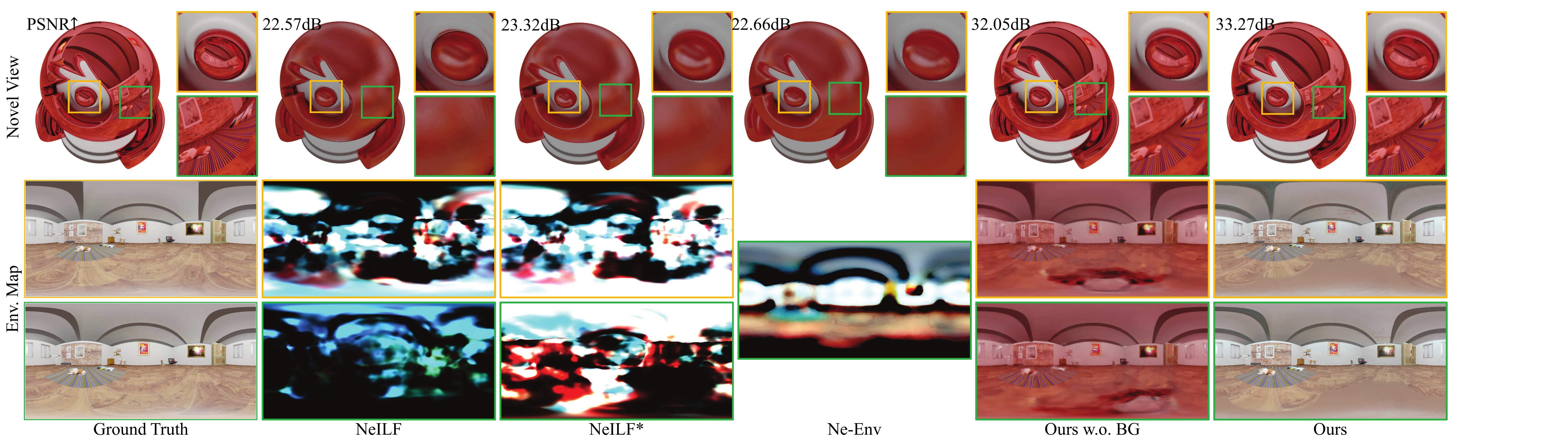}
    \vspace{-0.2in}
    \caption{Qualitative comparison of our method against baseline NeILF \cite{yao2022neilf} and its variants on self-rendered Glossy Blender dataset. The zoom-in details labeling with yellow and green boxes and corresponding reconstructed environment map surrounding the surface point of each box center are shown. Our method significantly outperforms other methods, both in the rendering quality of novel views and the reconstruction of ambient illumination. Noted that, although ours w/o background obtains reconstructs geometrically correct results, but fails on the decomposition of diffuse color and ambient illumination due to the incorrect white balance and less background supervision.
    }
    \vspace{-0.1in}
    \label{fig:cmp-blender}
\end{figure*}

\subsection{Loss} 
To alleviate the ambiguity of the material and illumination, we constrain the roughness $\rho(\mathbf{x}_s)$ and specular tint $\alpha(\mathbf{x}_s)$ of the surface point $\mathbf{x}_s$ to be relatively smooth. With the guidance of the image gradient of pixel $\mathbf{p}$, we defined the Bilateral Smoothness regularization \cite{yao2022neilf} as: 
\begin{equation}\small
    l_{s}\!=\!
    \frac{1}{|S_I|}
    \sum_{\mathbf{p} \in {S_I}}
        \left(\left \|  \nabla_{\mathbf{x}_s}\alpha(\mathbf{x}_s) \right\| \!+\!
         \left \|  \nabla_{\mathbf{x}_s} \rho(\mathbf{x}_s) \right\|\right)
         \mathbf{e}^
            { - \left \|  \nabla_\mathbf{p}\mathbf{I(p)} \right\|},
    \label{eq:smoothness}
\end{equation}
which forces the material gradient of the surface point $\mathbf{x}_s$ and its projected image pixel $\mathbf{p}$ to be corresponding.  The image gradient $\left \|  \nabla_\mathbf{p}\mathbf{I(p)} \right\|$ is pre-computed, and ${S_I}$ is the set of the pixel on the object.

We compute the L2-norm reconstruction loss between the predicted color $\widehat{\mathbf{C}}$ and the ground truth color $\mathbf{C}$ to jointly optimize the ambient illumination and object's materials:
\begin{equation}\small
    l_{rec}=\sum_{\mathbf{p} \in {S_I}} \left \| \widehat{\mathbf{C}}(\mathbf{x}_s,\mathbf{d}) - \mathbf{C}(\mathbf{p})  \right \|^2_2. 
    \label{equ:l2-loss}
\end{equation}
The regularization $l_{pre}$ of the pre-convoluted representation is performed in the same manner as the Eq. \eqref{equ:l2-loss}, where the pixel $\mathbf{p}$ is replaced with a pixel from the pre-convoluted background $S_B$, rather than from the object $S_I$. 
Similar to the hierarchical sampling procedure in NeRF, the proposed method also uses ``coarse" and ``fine" networks for further promising results and sampling efficiency. Overall, the entir loss in our method is $l=l_{rec}+\lambda_s l_{s}+\lambda_p l_{pre}$, where the weights $\lambda_s$  and $\lambda_p$ are empirically set to $10^{-4}$ and $10^{-1}$ in all our experiments. 



\section{Experiment}
\label{sec:exp}
\noindent\textbf{Implementation Details. }
Our method is implemented on top of Mip-NeRF \cite{barron2021mip} with PyTorch, and we discretize the Eq. (\ref{eq:our_render}) as Mip-NeRF. The number of samples for both the "coarse" and "fine" phases is 128.  We use the same architecture as Mip-NeRF (8 layers, 256 hidden units, ReLU activations) to train our ambient MLP network, but we apply the ILE module to featurize the input coordinates of incoming rays. We also use an 8-layer MLP with a feature size of 512 and a skip connection in the middle to represent the material MLP network. Please refer to our \emph{supplement} for more details about network setting and training schemes.

Through evaluating the results of novel view synthesis, the performance of the decomposition and illumination quality will be measured. We report the image quality with three metrics: PSNR, SSIM and LPIPS \cite{zhang2018unreasonable}, on both synthetic and real-world datasets.

\par\noindent\textbf{Baselines.} We compare our method with the following  methods: 1) NeILF \cite{yao2022neilf} modeled by the Disney BRDF and incident light field of each surface point; 2) NeILF* extended from NeILF which replaces the Disney BRDF to ours; and 3) Ne-ENV extended from NeILF* that uses a neural environment map instead of the incident light field. These methods using different illumination models could demonstrate the performance of our NeAI. The baseline implementation details could be found in our \emph{supplement}.


\begin{table}[tb] 
\centering
\caption{Quantitative comparison of our method and its ablations against baseline NeILF \cite{yao2022neilf} and its variants on Glossy Blender dataset. Metrics are averaged over all scenes. Our method shows far superior performance in terms of rendering quality.}
\small
\begin{tabularx}{\linewidth}{@{}l|CCC}
\hline
                                 & PSNR$\uparrow$   & SSIM$\uparrow$    & LPIPS$\downarrow$    \\ \hline
 NeILF                           & 25.851 & 0.930   & 0.088          \\
 NeILF$^*$                           & 27.279   & 0.941  & 0.084         \\
Ne-Env                 & 25.971   & 0.935  & 0.086          \\
\hline
w/o background                   & 36.404   & 0.979  & 0.033           \\
w/o pre-convolution       & 37.748   & 0.981  & 0.032           \\
 w/o regularization          & \cc{1}38.442   & \cc{1}0.983  & \cc{1}0.029
 \\  
 Ours                            & \cc{2}37.985   & \cc{2}0.981  & \cc{2}0.031           \\\hline
\end{tabularx}\vspace{-0.2in}
\label{tab:quan_synth}
\end{table}

\par\noindent\textbf{Glossy Blender Dataset.} 
Although previous NeRF variants have proposed various datasets containing diverse materials, the objects are all synthesized under the environment map, which ideally ignores the ambient distance. It is unrealistic and unnatural that an image renders without background and neglects the 3D environment.  
Therefore, we propose a new dataset closer to the natural condition. There are 7 synthetic scenes rendered in Blender \cite{blender}, each containing one glossy object placed in a 3D simulated environment with natural illumination. In our setting, 390 views are rendered around the upper hemisphere of the object, with 180 for training, 10 for validation, and 200 for testing. We adopt the rendered depth and normal maps as the result of stage one to isolate the influence of geometry quality and create an ideal test for evaluating lighting reconstruction and material decomposition.

\begin{figure*}[tb]
    \centering
    \includegraphics[width=1.0\linewidth]{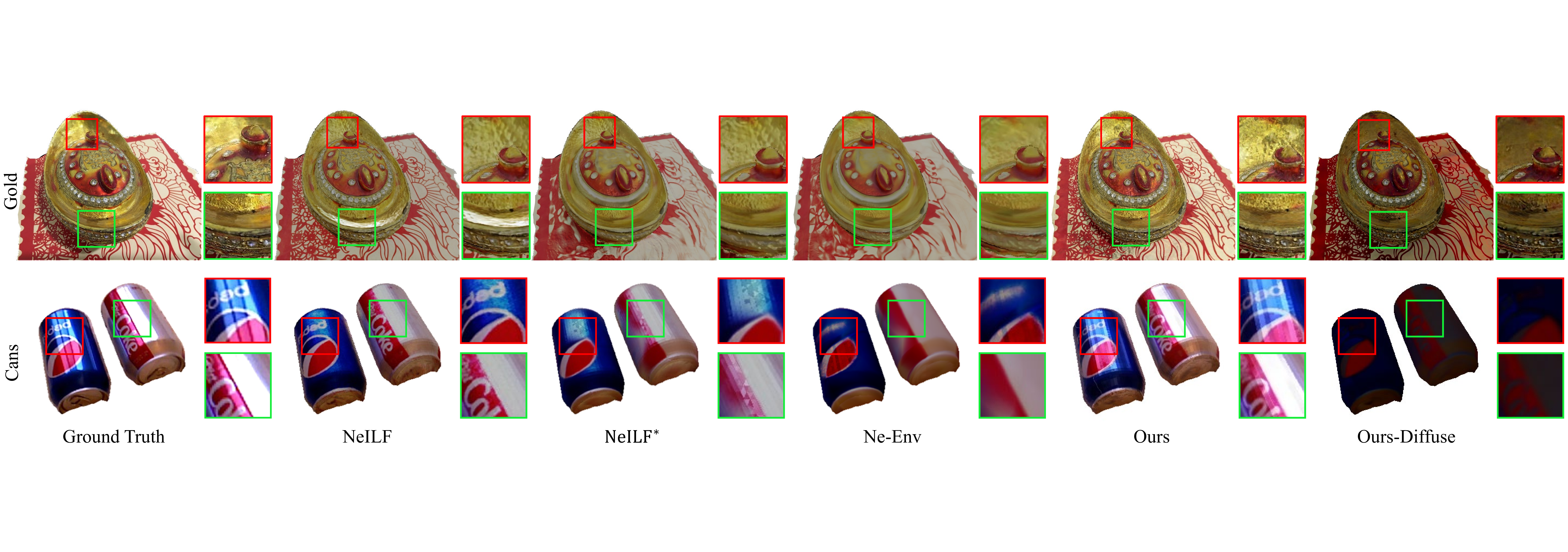}
    \vspace{-0.2in}
    \caption{Qualitative comparison with zoom-in details of our method against baselines on real-world \cite{yao2020blendedmvs, park2020seeing} dataset. Our method produces the most visually pleasing novel views, especially on reconstructing the highlight regions and detailed textures, as shown in zoom-in details. Note that the decomposed diffuse results of our method are also presented, where the high light is successfully removed. 
    }
    \vspace{-0.2in}
    \label{fig:cmp-blendedMVS}
\end{figure*}

Tab. \ref{tab:quan_synth} and Fig. \ref{fig:cmp-blender} demonstrate quantitative and qualitative comparisons of our method against baseline methods, respectively. Metrics in Tab. \ref{tab:quan_synth} are averaged over all scenes while the full experiments are accessible in \emph{supplement}. Considering the inputs of NeILF \cite{yao2022neilf} is the masked images of the objects without background, we test our method on the same setting for fairness, referring to our w/o background in the experiments. Besides, another two ablations are contained:
w/o pre-convolution, which does not take multiscale pre-convoluted representation into account; w/o regularization, which excludes the Bilateral smoothness regularization (Eq. \eqref{eq:smoothness}).
Tab. \ref{tab:quan_synth} and Fig. \ref{fig:cmp-blender} show the significantly superior performance of our method compared to baselines in terms of rendering quality on novel views, even without the guidance of background (w/o background). 

In particular, as for NeILF and NeILF$^*$, the derived environment maps indicates that they can't handle the nearby illumination, which changes dramatically with the position. Another reason is that they don't gather the information across different views, which is added to the ambiguity. Despite NeILF$^*$ simplifies the BRDF function, the same as our method, the performance compared with NeILF is slightly improved, but still has a large gap with our method, as shown in Fig. \ref{fig:cmp-blender} and Tab. \ref{tab:quan_synth}. It also verifies the performance of our NeAI on ambient illumination decomposition, especially the nearby lighting and complex 3D environments. As for Ne-Env, we share the same environment map across different positions. That reduces the uncertainty of the solution and therefore constructs a more meaningful environment map. However, the decomposition of materials and ambient illumination is poor. The reason is the simplification of the illumination model, \ie, all radiance incident upon the object being shaded comes from an infinite distance. 
Compared to NeILF and its variants, our method could render photo-realistic views and recover precise ambient illumination. Even without the background guidance, our method still reconstructs geometrically correct results. However, the white balance is not correct due to the majority color of the object is red, which causes ambiguity about the red object or the red incident light. This side-fact indicates it's necessary to use the background as guidance to disentangle the material and light.

\begin{table}[]\centering
\caption{Quantitative comparison of our method against baselines on Real-world Datasets. The table shows PSNR metrics of novel view renderings of test images. Our method consistently outperforms the other methods in terms of rendering quality.}\label{tab:quan_blendedMVS}
\vspace{-0.1in}
\scriptsize                     
\begin{tabularx}{\linewidth}{@{}l|CCCCCCCC}
\hline

                                 & Vase      & Cam     & Stone    & Gold    & Conch   & Bull   & Cans   & chips  \\ 
\hline
 NeILF                           &  \cc{2}26.270   & 17.182  & \cc{2}23.607   & \cc{2}18.458  & 19.327  & \cc{2}19.437   &\cc{2} 26.408  &\cc{2} 22.630  \\
 NeILF$^*$                        &  25.225   & \cc{2}19.205  & 22.269  & 17.369   & 19.272 & 19.337  & 25.565  & 21.571 \\
 Ne-Env                        &  24.564   & 18.668  & 21.378  & 17.511  & \cc{2}19.347  & 19.354   &23.560  & 21.906 \\
 Ours                            &  \cc{1}{28.927}     & \cc{1}{21.864} & \cc{1}{24.473}  & \cc{1}{20.549}  & \cc{1}{20.557}  & \cc{1}{22.865} &\cc{1} 31.574  & \cc{1}29.366  \\
\hline
\end{tabularx}\vspace{-0.1in}
\end{table}

\par\noindent\textbf{Real-world.} We then test our method on 8 real-world scenes from BlendedMVS \cite{yao2020blendedmvs} and Bag of Chips \cite{park2020seeing}, which provide images and depth maps reconstructed by MVS methods or RGBD camera. 
We selected the last ten images as the test set for BlendedMVS, while for Bag Of Chips, we left the last 1/5 as the test set.
The qualitative and quantitative comparisons are shown in Tab. \ref{tab:quan_blendedMVS} and Fig. \ref{fig:cmp-blendedMVS}  respectively. 
All the results validate the performance of our method on illumination reconstruction and material decomposition, representing the robustness of our method in geometric perturbation.
Compared to baselines, the performance gap degrades for two reasons: 1) the images with varying exposures and rough geometry in real-world datasets make the decomposition difficult; 2) not enough background in original images could be used to help the ambient illumination reconstruction.

\begin{figure}[tb]
    \centering
    \includegraphics[width=1\linewidth, trim=0 1.4cm 0 0]{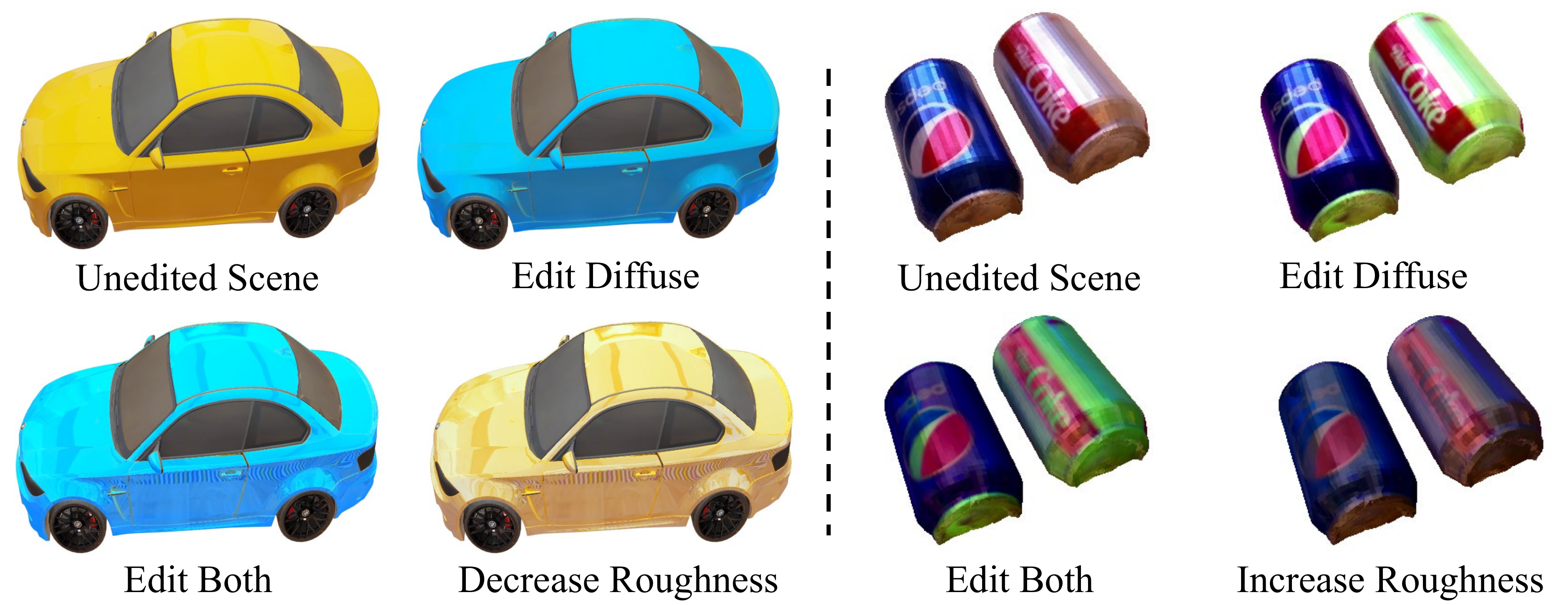}
    \caption{Our model decomposes the roughness and diffuse color from ambient lighting that enables editing. We can manipulate the roughness of by multiplying a global scale, which demonstrates the specular reflections naturally. We can also edit the diffuse color without affecting its glossy paint's specular reflections.}
    \vspace{-0.25in}
    \label{fig:app_mtl}
\end{figure}

\begin{figure*}[th]
    \centering
    \includegraphics[width=1.0\linewidth]{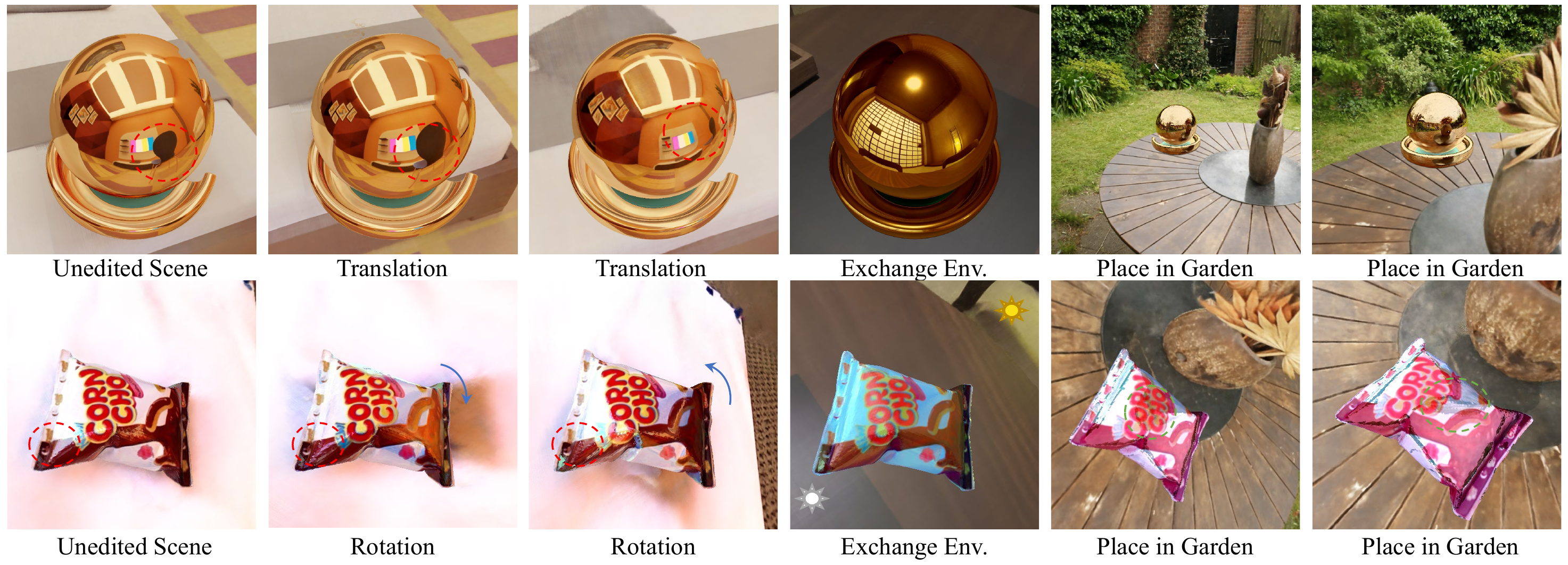}
    \vspace{-0.25in}
    \caption{Visualizations of illumination manipulations. Our method recovers environment lighting that enables novel view rendering with high-fidelity specularities and natural illumination under different manipulations. Our method naturally handles the occlusion of a chair in translation cases and the high lighting on the cornchos in rotation cases. We place our decomposed objects in a virtual environment or pre-trained Mip-NeRF environment and produce novel views with more realistic reflections ({\bf Better viewed on screen with zooming in}).
    }
    \vspace{-0.1in}
    \label{fig:app-env}
\end{figure*}

\begin{figure}[t]
    \centering
    \includegraphics[width=1.\linewidth]{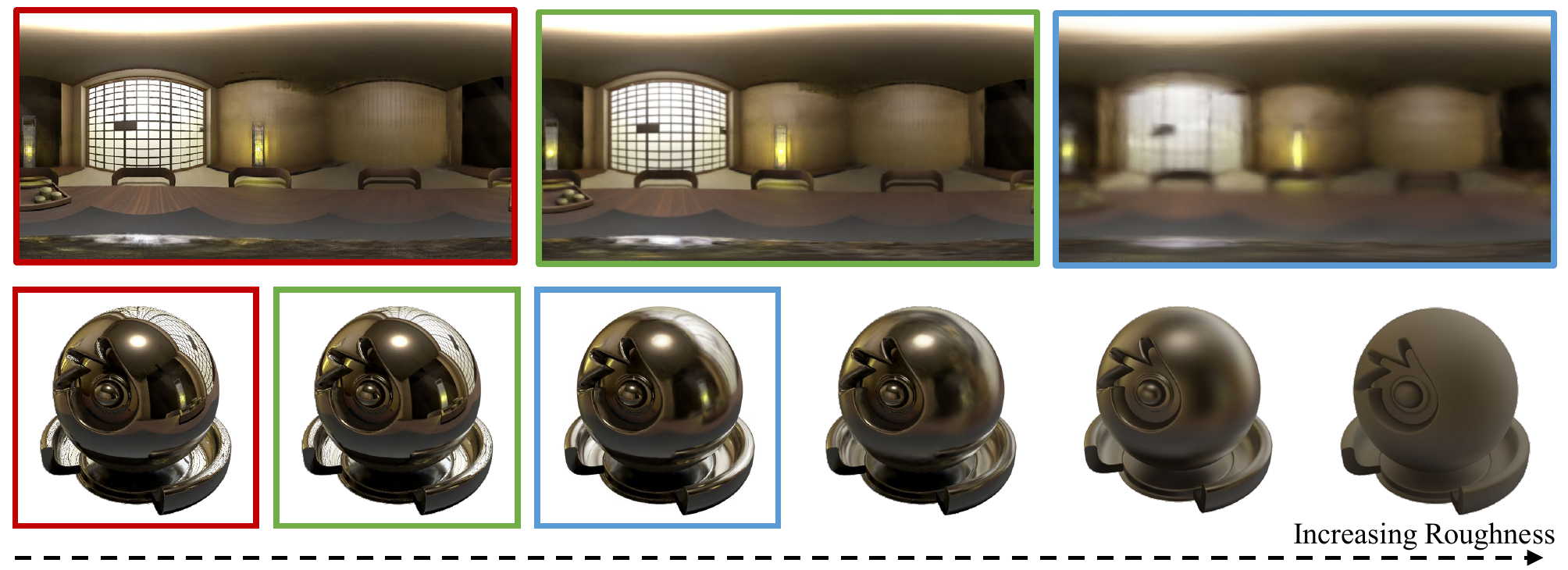}
   \vspace{-0.2in}
    \caption{{Visualizations of the roughness editing. 
    The first row shows the corresponding environment maps for the first three balls, and the second row shows that the ball gradually becomes blurred with increasing roughness. Noted that, although the roughness is beyond the Gaussian kernel used in the pre-convoluted representation, our method could also render visually realistic results.}}
    \vspace{-0.2in}
    \label{fig:app_mtl_sphere}
\end{figure}

\section{Applications} 

\noindent\textbf{Object's material editing.} As our model disentangles the object's material well, our components behave intuitively and enable visually reasonable material editing results. Fig. \ref{fig:app_mtl} shows material decomposition (\eg roughness and diffuse color) and natural example edits of the object's materials on the `Car' dataset. 
Furthermore, Fig. \ref{fig:app_mtl_sphere} shows convincing results of roughness editing
(second row) and their corresponding environment maps (first row) on the 'Metal Ball' dataset. As the roughness increases, novel views gradually become blurred, even when the roughness becomes extremely large that exceeds the maximum scale of pre-convoluted representation. 

\par\noindent\textbf{Illumination manipulation.} 
Fig. \ref{fig:app-env} illustrates several types (rotation, translation, exchange) of manipulation to the ambient illumination around the object. Our method produces natural photo-realistic views, especially for the environment occlusions (\eg reflected chair) and highlights (\eg chips) as shown in the translation and rotation cases respectively. More importantly, with a slight scale to the decomposed materials, we place our plug-and-play NeAI model in a pre-trained Mip-NeRF environment \cite{barron2022mip}, as shown in the last two columns of Fig. \ref{fig:app-env}. Although there is complex and detailed illumination, visually harmonious novel views are re-rendered with more realistic reflections in the ``Garden'' dataset.
It verifies that our NeAI is an easy-to-use plugin for NeRF that gives objects a better sense of belonging in the new 3D NeRF-style environments. Please refer to our supplement video for better visualization.


\section{Conclusion}
\label{sec:con}
This paper presents a novel neural approach to efficiently and stereographically modeling 3D ambient illumination. Previous methods focus on simplified lighting models (\eg environment map and spherical Gaussian) to represent no-distant illumination. Instead, we propose NeAI to model illumination as volumetric radiance fields such that each sample of the surrounding 3D environments is equivalent to a light emitter. We show that, together with our integral lobe encoding and pre-convoluted representation, our method can accurately recover ambient illumination and naturally re-render high-quality views for a decomposed object under new NeRF-style environments.
We believe that with this high-fidelity and fully differentiable lighting representation, it can be easily extended to downstream tasks and bring us closer to bridging the gap between virtual and real-world scenes.

\noindent\textbf{Limitations. } {It is difficult to model ambient illumination and decompose the object's material, our approach has the following limitations. First, our pipeline relies on the geometry reconstructed through stage one, similar to \cite{yao2022neilf, zhang2021nerfactor}. 
Second, most illumination is HDR for better harmonization. Although we use a gamma correction to approximate the LDR-HDR mapping, our method could use a similar tone-mapping module in HDR-NeRF \cite{huang2022hdr} to recover HDR-NeAI.} 
We leave those problems in future works.

\clearpage
\setcounter{section}{0}

\renewcommand\thesection{\Alph{section}}
\renewcommand\thesubsection{\Alph{section}.\arabic{subsection}}

\section*{\Large \centering Supplementary Material}   

\section{Overview}
The supplementary material provides the derivation of integral lobe encoding (ILE, Eq. (6) in the submission) (Section \ref{sec:derivation}), the detailed architecture of our method, and the implementation details of the baseline methods (Section \ref{sec:implement}). We also introduce more details of our plug-and-play rendering process. The additional experiments demonstrate the superior performance of our method compared to the baseline methods (Section \ref{sec:experiment}). We \textbf{strongly} encourage the reader to see our website-video supplementary \gitlink, in which we present our results on various applications, especially our plug and play: place meshes in the public 360v2 datasets\footnote{\href{https://jonbarron.info/mipnerf360/}{https://jonbarron.info/mipnerf360/}} \cite{barron2022mip}. 

\section{Derivation of Integral Lobe Encoding}
\label{sec:derivation}


Here, we introduce our ILE featurization procedure, in which we cast a lobe whose size is defined by the spatially-varying roughness. We feature conical frustums along that roughness-varying lobe. Similar to Mip-NeRF \cite{barron2021mip}, images in our method are rendered one pixel at a time so that we can describe our procedure in terms of an individual pixel of interest being rendered. Mip-NeRF casts a constant cone from the camera's center along the view direction that passes through the pixel's center, where the cone's size is spatially-constant and defined by the width of the pixel in world coordinates. As shown in Eq. (2) of the submission, we instead decompose the pixel's color into diffuse color and specular color of surface point $\mathbf{x}_s$, where $\mathbf{x}_s$ is the intersection between the object surface and the ray from the camera's center along the view direction. As shown in Eq. (5) of the submission, the specular color is the integration of the radiance field of each sampled point within the specular lobe combined with surface material. The apex of that specular lobe lies at $\mathbf{x}_s$, and the radius of the specular lobe is defined by $\hat{\rho}\hat{r}$. According to the pre-filtered environment map used in traditional CG rendering \cite{akenine2019real}, we set $\hat{\rho}$ to the material roughness $[0, 1]$ scaled by the pixel's size $r_0$, and set $\hat{r}$ to the width of the pixel $r_0$ scaled by mipmap levels (\ie resolution multi-scaling). In the implementation, we discretely divide the specular lobe into several conical frustums according to the discrete distance.
The set of positions $\mathbf{x}$ that lie within a conical frustum between two distant value $[t_0, t_1]$ is then defined as,
\begin{equation}\small
 \begin{split}
    F(\mathbf{x}, \mathbf{x}_s,\mathbf{l}_r,\rho,\hat{r},t_0,t_1)=\mathfrak{l}\left\{\left(t_0<\frac{\mathbf{l}_r^\top(\mathbf{x}-\mathbf{x}_s)}{\left\|\mathbf{l}_r\right\|^2_2}<t_1\right)\right.\\
    \wedge\left.\left(\frac{\mathbf{l}_r^\top(\mathbf{x}-\mathbf{x}_s)}{\left\|\mathbf{l}_r\right\|_2\left\|\mathbf{x}-\mathbf{x}_s\right\|_2}>\frac{1}{\sqrt{1+\left(\hat{\rho}\hat{r}/\|\mathbf{l}_r\|_2\right)^2}}\right)\right\},
\end{split}
\label{eq:concial_frustum}
\end{equation}
where $\mathfrak{l}\left\{\cdot\right\}$ is an indicator function: $F(\mathbf{x}, \cdot) = 1$ iff $\mathbf{x}$ is within the conical frustum defined by $(\mathbf{x}_s,\mathbf{l}_r,\rho,\hat{r},t_0,t_1)$. $\|\cdot\|_2$ refers to the Euclidean norm of a vector. Note that the conical frustum's radius is defined not only by the pixel width but also by the roughness, which means the roughness-varying specular lobe instead of constant.

Besides, the success of NeRF in high-fidelity rendering results is attributed to the proposed position encoding for learning high-frequency information. Each sample point $\mathbf{x}$ goes through this featurization procedure before being fed into the MLP, which is formulated as, 
 \begin{equation}\small
    \mathrm{PE}(\mathbf{x}) \!=\! \left[  \sin(\mathbf{x}),\cos(\mathbf{x}),\ldots,\sin(2^{L-1}\mathbf{x}), \cos(2^{L-1}\mathbf{x})\right]^T,
    \label{eq:pe}
 \end{equation}
where $L$ is a hyperparameter and determines the bandwidth of the interpolation kernel. This encoding enables the MLP parameterizing the scene to behave as an interpolation function, and maps the position into a high-frequency space. Consequently, we establish a featurized representation of Eq. \eqref{eq:pe} inside the conical frustum defined by Eq. \eqref{eq:concial_frustum},
\begin{equation}\small
 \begin{split}
  \mathrm{ILE}(\mathbf{x}_s,\mathbf{l}_r, p_l)
  = \frac{\int A(\mathbf{x}, \mathbf{x}_s,\mathbf{l}_r) \mathrm{PE}(\mathbf{x}) {F}(\mathbf{x}, \mathbf{x}_s,\mathbf{l}_r,p_l)\mathrm{d}\mathbf{x}}
  {\int {F}(\mathbf{x}, \mathbf{x}_s,\mathbf{l}_r,p_l)\mathrm{d}\mathbf{x}},\\
  \mathrm{where}\ A(\mathbf{x}, \mathbf{x}_s,\mathbf{l}_r) = \exp{\left(\frac{\mathbf{l}_r^\top(\mathbf{x}-\mathbf{x}_s)}{\left\|\mathbf{l}_r\right\|_2\left\|\mathbf{x}-\mathbf{x}_s\right\|_2}-1\right)}
    \end{split}
    \label{eq:ILE_Int}
\end{equation}
where $p_l = \{\hat{\rho},\hat{r},t_0,t_1\}$ indicates the shape parameters about the conical frustum. According to Eq. (5) in the submission, a larger roughness corresponds to a rougher surface under a wider vMF distribution, and the integral volume is attenuated with the angle between incoming direction $\mathbf{l}={\mathbf{x}-\mathbf{x}_s}$ and the reflection direction $\mathbf{l}_r$ of the ambient illumination. Consequently, we consider the ILE must be under a vMF distribution, and it attenuates with the weights of $\mathbf{l}_r^\top(\mathbf{x}-\mathbf{x}_s)$. We then set the $A(\mathbf{x}, \mathbf{x}_s,\mathbf{l}_r)$ to the attenuation function, which can be well-approximated using a simple exponential function. Considering attenuation function $A$ is related to the direction of incoming illumination under a vMF distribution, the integral in the numerator can be defined via a multivariate Gaussian to efficiently represent the vMF attenuation and compute that features one time. 
According to integral position encoding (IPE) in Mip-NeRF \cite{barron2021mip}, the integration within a conical frustum in the numerator has no closed-form solution, only Gaussian approximation. However, this approximation will be inaccurate if there is a significant difference between the center and the boundary of the frustum. Compared to IPE, our ILE is the vMF attenuation, which is more suitable and accurate to describe as a multivariate Gaussian expression.

Eq. \eqref{eq:ILE_Int} can be directly simplified as a multivariate Gaussian. We first represent the shape of the conical frustum by the mean distance along the ray $\mu_t$, the variance along the ray $\sigma_t^2$, and the variance perpendicular to the ray $\sigma_r^2$. According to a midpoint $t_\mu=(t_0+t_1)/2$ and a half-width $t_\sigma=(t_1-t_0)/2$, these variables can be parameterized as,
 \begin{equation}\small
 \begin{split}
  \mu_t=t_\mu+\frac{2 t_\mu t_\delta^2}{3 t_\mu^2 + t_\delta^2},
  \sigma_t^2 = \frac{t_\sigma^2}{3} - \frac{4t_\delta^4(12t_\mu^2 - t_\sigma^2)}{15 (3 t_\mu^2+t_\delta^2)^2},
  \\
  \sigma^2_r(\hat{\rho}) = \hat{\rho}^2\hat{r}^2\left(\frac{t_\mu^2}{4}+ \frac{5t_\sigma^2}{12} - \frac{4t_\sigma^4}{15(3t_\mu^2+t_\delta^2)} \right),
  \end{split}
  \label{eq:cf_gaussian}
 \end{equation}
where the width of conical frustum $\sigma_r^2(\rho)$ is spatially varied with the roughness of surface point. 

We then compute the shape of a set of conical frustums along the specular lobe. We transform Eq. \eqref{eq:cf_gaussian} from the coordinate frame of conical frustums into the world coordinates, of which the mid-point $\mathbf{\mu}$ and variance $\mathbf{\Sigma}$ are defined as follows, 
\begin{equation}\small
    \mathbf{\mu}=\mathbf{x}_s+\mu_t\mathbf{l}_r,
    \mathbf{\Sigma}(\hat{\rho})=\sigma_t^2\left(\mathbf{l}_r\mathbf{l}_r^\top\right)+\sigma^2_r(\hat{\rho})\left(\mathbf{I}-\frac{\mathbf{l}_r\mathbf{l}_r^\top}{\left\|\mathbf{l}_r\right\|_2^2}\right),
    \label{eq:specular_lobe}
\end{equation}
where the variance $\mathbf{\Sigma}$ is related to spatially-varying roughness $\rho$. The next derivation is similar to Mip-NeRF \cite{barron2021mip}, Eq. \eqref{eq:pe} can be rewritten as matrix form ,
\begin{align}\small
    &\mathrm{PE}(\mathbf{x})=\left[\sin{(\mathbf{P}\mathbf{x}), \cos{(\mathbf{P}\mathbf{x})}}\right]^\top\\
    &\mathbf{P}\!\!=\!\!\left[
    \begin{matrix}
    1 & 0 & 0 & 2 & 0 & 0 & &2^{L-1} & 0 & 0\\
    0 & 1 & 0 & 0 & 2 & 0 & \cdots & 0 &2^{L-1} & 0 \\
    0 & 0 & 1 & 0 & 0 & 2 & & 0 & 0 &2^{L-1} \\
    \end{matrix}\right]^\top
    \label{eq:pe_mat}
\end{align}
This matrix form could help us to derive the final ILE feature. According to Eq. \eqref{eq:pe_mat}, we could identify the mean and covariance of the conical frustum Gaussian along the specular lobe. Our ILE feature can be finally formulated as the expected sines and cosines of the mean and the diagonal of the covariance matrix:
 \begin{equation}\small
 \begin{aligned}
\mathrm{ILE} ( \mathbf{x}_s,\mathbf{l}_r, p_l ) &= E_ {x\sim N(\mu,\mathbf{\Sigma}(\hat{\rho}))}[\mathrm{PE} (x)]
\\&= \left[ \begin{aligned}{\sin ( \mathbf{P}\mathbf{\mu} )\odot exp(-\frac{1}{2}\mathrm{diag}( \mathbf{P}\mathbf{\Sigma}(\hat{\rho})\mathbf{P}^T))}
\\{\cos ( \mathbf{P}\mathbf{\mu} )\odot exp(-\frac{1}{2}\mathrm{diag}( \mathbf{P}\mathbf{\Sigma}(\hat{\rho})\mathbf{P}^T))}   \end{aligned}\right],
 \end{aligned}
 \label{eq:ILE}
 \end{equation}
where $\odot$ refers to element-wise multiplication. For better understanding, we formulate Eq. \eqref{eq:ILE} as 
\begin{equation}\small
        \mathrm{ILE}(\mathbf{x}_s,\mathbf{l}_r, p_l) \!=\! \left\lbrace \!
        \begin{bmatrix}
            {\sin( 2^{\ell}\mathbf{\mu} )\exp(-2^{\ell-1}\mathrm{diag}(\mathbf{\Sigma}(\hat{\rho})))} \\
            {\cos ( 2^{\ell}\mathbf{\mu})\exp(-2^{\ell-1}\mathrm{diag}(\mathbf{\Sigma}(\hat{\rho})))}   
        \end{bmatrix}\!
        \right\rbrace_{\ell=0}^{L-1},
    \label{eq:ILE_final}
\end{equation}
\begin{equation}\small
    \mathrm{diag}(\mathbf{\Sigma}(\hat{\rho}))=\sigma_t^2\left(\mathbf{l}_r\odot\mathbf{l}_r\right)+\sigma^2_r(\hat{\rho})\left(\mathbf{1}-\frac{\mathbf{l}_r\odot\mathbf{l}_r}{\left\|\mathbf{l}_r\right\|_2^2}\right)
\end{equation}
where $L$ is a hyperparameter and determines the bandwidth of the interpolation kernel. $p_l = \{\hat{\rho},\hat{r},t_0,t_1\}$ refers to the shape parameters about the conical frustum.  Our ILE allows the MLP to parameterize the incoming illumination inside the roughness-varying specular lobe, whose strength is variable with the incoming direction under vMF distribution, to behave as an interpolation function. As a result, our ILE efficiently maps continuous input coordinates into a high-frequency space.

\begin{figure}[h]
\centering
\includegraphics[width=0.7\linewidth]{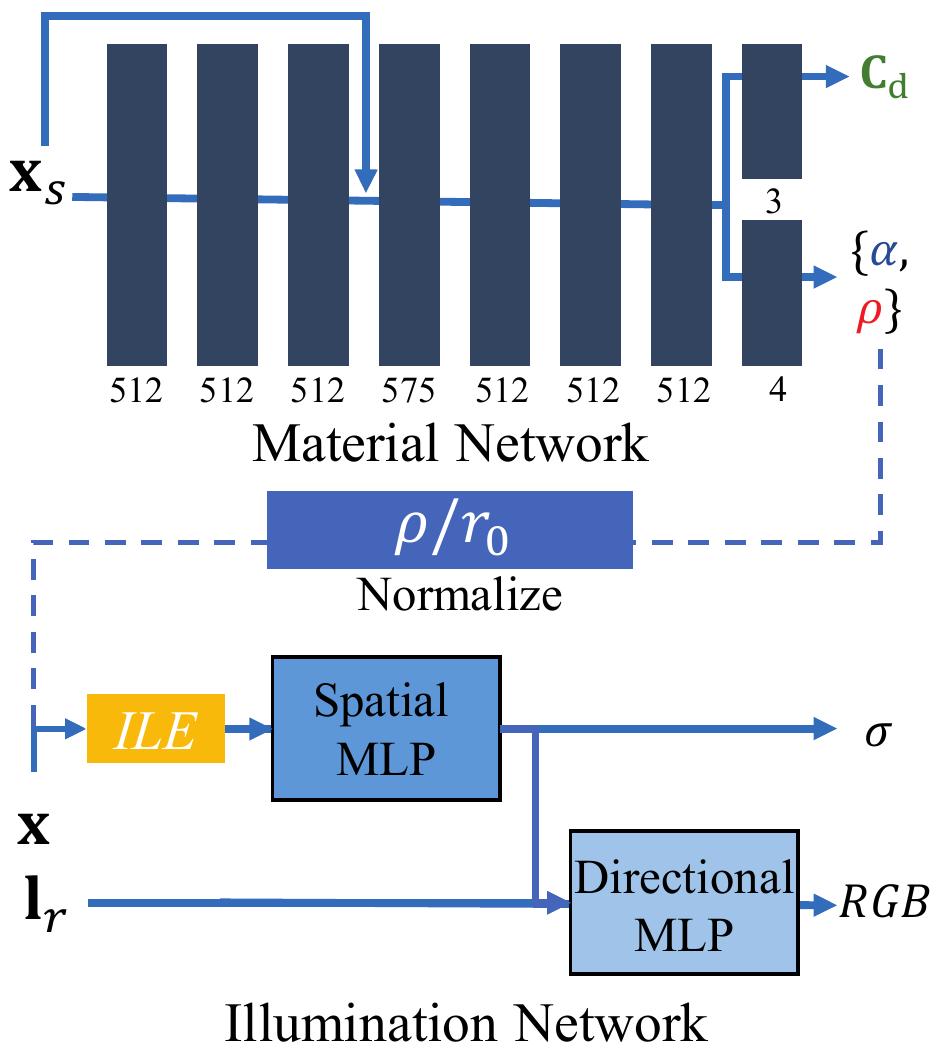}
\caption{Details of the network architecture. As for the material network, a ReLU is applied for each MLP except the last layer. We choose tanh and sigmoid to the output of $\mathbf{C}_d$ and ${\alpha, \rho}$, separately. The  parameters of illumination network is the same as \cite{barron2021mip}. }
\label{fig:network}
\end{figure}
\section{Implement Details}
\label{sec:implement}

\begin{figure}[th]
\centering
\includegraphics[width=\linewidth]{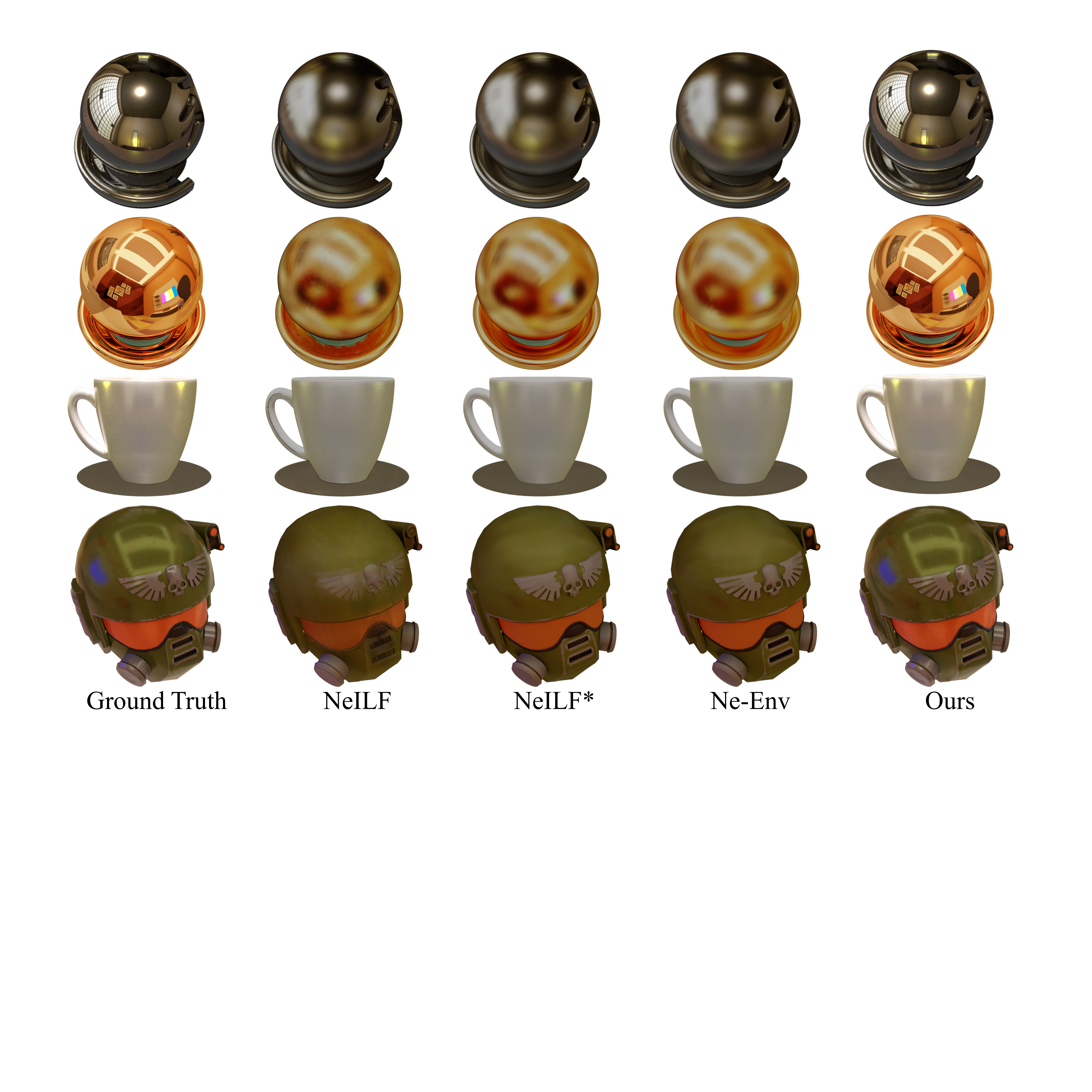}
\caption{Visualisations of the results on our "Glossy Blender" dataset. It' is obvious that the reflection and the highlights are correctly reconstructed with our method. Compared to NeILF or other baselines, it shows great robustness in representing complex lighting.
 } 
\label{fig:cmp_blender}
\end{figure}

\begin{figure*}[th]
\centering
\includegraphics[width=\linewidth]{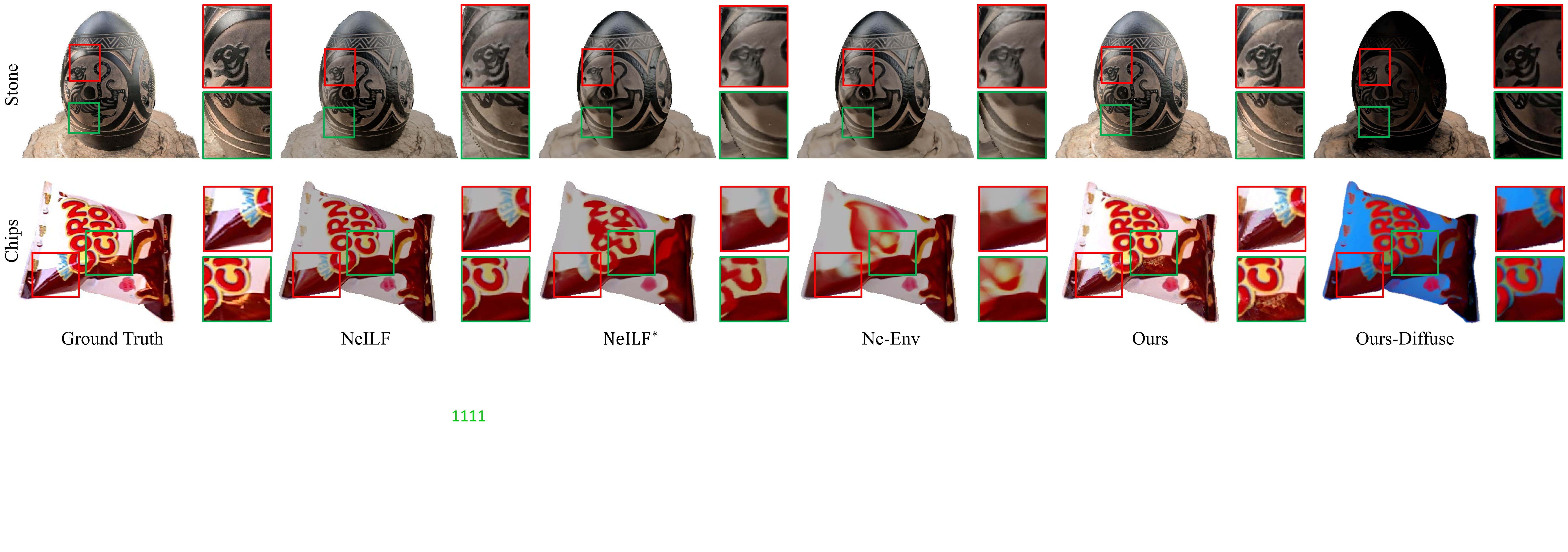}
\caption{More qualitative comparison with zoom-in details of our method against baselines on real-world \cite{yao2020blendedmvs, park2020seeing} dataset. Our method produces the most visually pleasing novel views, especially on reconstructing the highlight regions and detailed textures, as shown in zoom-in details. Note that the decomposed diffuse results of our method are also presented, where the highlights are successfully removed.  The diffuse color of our method looks wrong on white balance, probably because lacking of enough background information.
 }
\label{fig:cmp_real}
\end{figure*}

\subsection{Architecture}
In the paper, we introduce our novel representation of illumination, and the rendering equation for the pixel of the object is given as Eq. (7). The whole pipeline is differentiable, and the architecture of the network is shown in Fig. \ref{fig:network}. We believe the architecture can be easily re-implemented or extended by other researchers.

For the pixel of the background, we remove the diffuse term $\mathbf{C}_d$ and the tint $\alpha$, and directly give the value of radius $r_p=3\sigma r_0$. $\sigma$ is set according to the pre-convoluted image. It is worth noting that, different with traditional pre-filtered technique \cite{akenine2019real}, our method does not handle the images with different mipmap levels (multiscale resolution) for simplification, as shown in Fig. 3, so $\hat{r}$ is set to the raw pixel's size $r_0$.
According to Eq. (3) and ILE Eq. \eqref{eq:ILE_final}, the captured value of along the ray $\mathbf{r}(t)=\mathbf{o}+ t \mathbf{d}$ from $t_n$ to $t_f$ is:
\begin{equation}\small
\begin{aligned} \label{eq:rendering}
 \mathbf{C}(\mathbf{o},\mathbf{d}) &= \int_{t_n}^{t_f}{T(t)\sigma(\mathcal{L}})  \mathbf{C}(\mathcal{L},\mathbf{d})\mathrm{d}t ,
\\
 	\textup{where} \:
 	\mathcal{L} &= \mathrm{ILE}(\mathbf{r}(t),\mathbf{d},p_l),
\end{aligned}
\end{equation}
and $p_l = \{3\sigma,r_0,t_n,t_f\}$ refers to the shape parameters about the conical frustum along that ray.

\subsection{More about Plug and Play} With our novel representation of environment lighting and its interaction with the object surface, we lay the foundation for a fancy application: seamless illumination of a decomposed mesh with a NeRF-style environment.
 For a decomposed object from stage two, we need to manually set the transformation to put it into the environment.
Sequentially, the realistic results can be rendered through the rendering equations shown in Eq. (7) of the submission and Eq. \eqref{eq:rendering}. 
Considering that the object is sometimes occluded by the landscape, we modify the rendering Eq.(7) slightly to account for occlusions during the rendering of new views,
\begin{equation}\small
    \mathbf{C}(\mathbf{o}, \mathbf{d}) \!=\!  
        \mathbf{C^*}(\mathbf{o},\mathbf{d})  w^*(\mathbf{o},\mathbf{d}) 
        + \mathbf{C}( \mathbf{x}_s,  \mathbf{d}) (1-w^*(\mathbf{o},\mathbf{d})),
\end{equation}
where $\mathbf{C^*(\cdot)}$ and $w^*(\cdot)$ with superscript * mean that the $t_f$ is set to $0.99$ times the depth of the object's surface. $w^*(\cdot)$ is the cumulative distribution function of density from $t_n$ to $t_f$ along the ray. Additionally, it is recommended to scale the decomposed material $\alpha,\rho,\mathbf{C}_d$ for better visual effect. The rendering videos could be found in \textit{supplement website}.

It is worth noting that, with our rendering pipeline, our method takes about 50 seconds to render an $800\times800$ image on a single Tesla A100, while the Cycles engine on Blender \cite{blender} takes about several minutes to render the ground truth.

\subsection{Traning Details}
Our implementation is based on the re-implementation pytorch \cite{Pytorch} of Mip-NeRF \cite{barron2021mip}.
We train our model with Adam \cite{kingma2015adam} for 1 million iterations.
For each iteration, 4096 rays are randomly selected from all images including the pre-convoluted images. The training process takes about two days to train on a single Tesla A100 GPU or 16 hours for 600,000 iterations on 4 Tesla A100 GPUs.

\subsection{Baseline Implementations}
\noindent\textbf{NeILF. \cite{yao2022neilf}} We follow the same setup in the author's code\footnote{\href{https://github.com/apple/ml-neilf}{https://github.com/apple/ml-neilf}}. The architecture can be divided into two parts, one for material estimation, called BRDF MLP, and the other, NEILF MLP. We follow their procedure for training on a single scene.

\noindent\textbf{NeILF$^*$.} We replace the BRDF MLP with our material network and modify the rendering equation to directly predict the diffuse term as $\mathbf{C}_d$. Another change is to modify the Fresnel term F to be a position-dependent magnitude $\alpha$, so the specular term is only related to roughness $\rho$, magnitude $\alpha$. The predicted $\mathbf{C}_s$ is finally obtained by integrating with the product of all incident light and the specular term on the hemisphere $\Omega$.

\noindent\textbf{Ne-Env.} A variant of NeILF$^*$, named neural environment map. The position point input of NEILF MLP is removed, so the spatially-varying environment map is modified to the non-spatially varying environment map.

\noindent\textbf{w/o ILE.} The roughness $\rho$ is set to a small constant as $r_0$, which means the adaptive lobe encoding is abolished, and the specular lobe degenerates into a thin ray. It causes artifacts to the objects with rough material. 

\section{Additional Results}
\label{sec:experiment}
\subsection{Glossy Blender dataset}

The error metrics for each individual scene are provided in Tabs. \ref{tab:quan_glossy_psnr}, \ref{tab:quan_glossy_ssmi} and \ref{tab:quan_glossy_lpips}, and the more visualized results are shown in Fig. \ref{fig:cmp_blender}. It demonstrates the significantly superior performance of our lighting representation compared to the baselines through rendering quality on novel views.  

\begin{figure}[th]
\centering
\includegraphics[width=\linewidth]{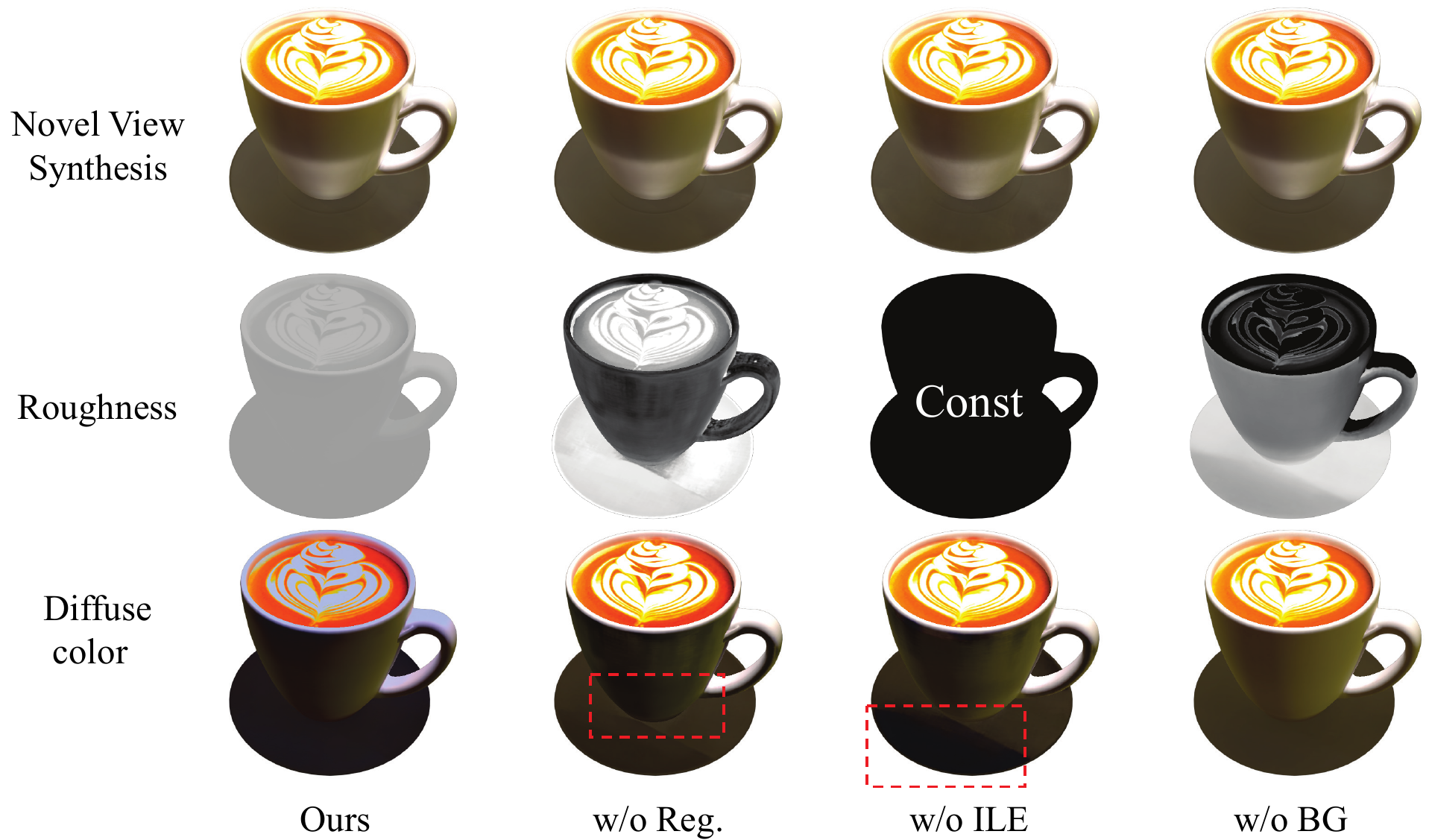}
\caption{Visualizations of the decomposed material for ablation study. Without the Bilateral Smoothness regularization, the material becomes noisy. Without the ILE, changes in material roughness cannot be handled well, resulting in artifacts of diffuse items. Without background, the roughness is not correct due to the effects of lighting.
 }

\label{fig:ablation}
\end{figure}

We also show visual comparison of the ablation in Fig. \ref{fig:ablation}. We can see that, without the Bilateral Smoothness regularization (Eq. (8) in our submission), although obtains the best metric performance, it generates noise and becomes inconsistent across the same material. It is unsuitable for material editing applications. On the other hand, without the ILE, spatially-varying roughness can not be handled very well, resulting in artifacts of diffuse items, as shown in Fig. \ref{fig:ablation}. The result of 'w/o BG' shows that without the assistance of the background, it's easy to creates ambiguity between light and material.


\subsection{Real-world dataset}

Tabs. \ref{tab:quan_synth_ssim} and \ref{tab:quan_synth_lpips} list the average SSIM and LPIPS for each individual scene in real-world dataset \cite{yao2020blendedmvs, park2020seeing}, and more visulization is shown in Fig. \ref{fig:cmp_real}. It demonstrates the performance of our method in real scenes compared to NeILF and its variants in terms of rendering quality on novel views. 


\begin{table}[tb] 
\centering
\caption{Per-Scene quantitative comparison of SSIMs on real-world dataset.}
\scriptsize 
\begin{tabularx}{\linewidth}{@{}l|CCCCCCCC}
\hline
                   &bull      &cam       &stone       &gold         &conch          & vase                          &cans    &chips\\ \hline
 NeILF             &\cc{1}0.779     &0.460     &\cc{2}0.859      &\cc{1}0.771       &\cc{1}0.787  &\cc{2}0.827      &\cc{2}0.957  &\cc{2}0.939             \\
 NeILF$^*$         &0.725         &\cc{2}0.573     &0.801        &0.625          &0.744           &0.745             &0.949  &0.917   \\
Ne-Env             &0.745         &0.567     &0.801              &0.633           &0.750         &0.738             &0.944  &0.915  \\
 Ours               &\cc{2}0.760   &\cc{1}0.705   & \cc{1}0.862   &\cc{2}0.769     &\cc{2}0.655     &\cc{1}0.851     &\cc{1}0.960  &\cc{1}0.953      \\\hline
\end{tabularx}
\label{tab:quan_synth_ssim}
\end{table}

\begin{table}[thb] 
\centering
\caption{Per-Scene quantitative comparison of LPIPS on real-world dataset.}
\scriptsize 
\begin{tabularx}{\linewidth}{@{}l|CCCCCCCC}
\hline
                   &bull      &cam       &stone       &gold      &conch     & vase                                   &cans    &chips\\ \hline
 NeILF             &\cc{2}0.272     &0.437    &\cc{2}0.131      &\cc{1}0.173      &\cc{1}0.313     &\cc{2}0.213      &\cc{2}0.034  &\cc{2}0.051            \\
 NeILF$^*$         &0.318        &\cc{2}0.345  &0.172           &0.297             &0.382     &0.301                 &0.044  &0.076 \\
Ne-Env             &0.324        &0.355        &0.187           &0.311              &0.391     &0.311                 &0.050  &0.082 \\
 Ours               &\cc{1}0.257   &\cc{1}0.256   &\cc{1}0.130      &\cc{2}0.186        &\cc{2}0.366     &\cc{1}0.172    &\cc{1}0.028  &\cc{1}0.041       \\\hline
\end{tabularx}
\label{tab:quan_synth_lpips}
\end{table}

\begin{figure}[th]
\centering
\includegraphics[width=1\linewidth]{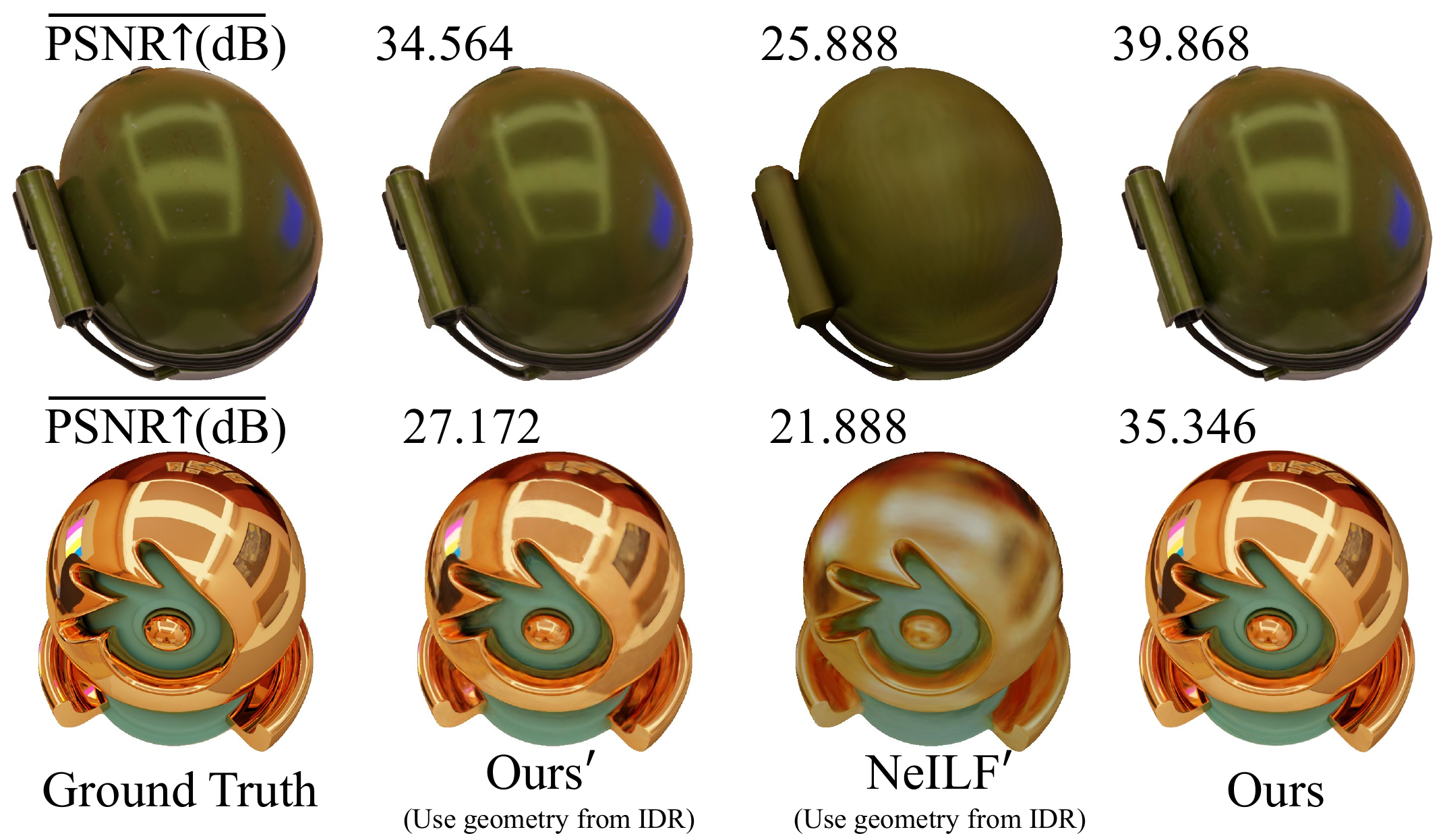}
\caption{Qualitative and quantitative results based on a geometry reconstructed by IDR\cite{yariv2020multiview} (in column 2,3). Compared to the result of the ideal test (in column 4), it is worth noting that the geometric deviation from the stage one would affects the quality of novel-view synthesis. This is because the tracing direction and origin of the specular lobe depends on the geometry. For a rougher surface, the lobe is wider, so is has more tolerance to this perturbation. The opposite is true for a smoother surface. We demonstrate this effect with two examples of different roughness.
 }

\label{fig:geometry_disc}
\end{figure}

\begin{table*}[t]
\centering
\caption{Per-Scene quantitative comparison of PSNRs on Glossy Blender dataset.}
\small
\begin{tabularx}{\linewidth}{@{}l|CCCCCCC}
\hline
                   & Car  &Helmet & Coffee  &RedBall    & GoldBall & GoldBall & MetalBall  \\ 
                   & Playroom  &Bedroom & Tearoom  &Playroom    & Bedroom & Bedroom2 & Tearoom  \\
\hline
 NeILF             & 20.863   & 28.282    &26.785      &26.630         &24.889         &24.620        &28.886        \\
 NeILF$^*$         & 21.332   & 29.419    &27.036      &27.444         &25.383         &25.175        &29.216  \\
Ne-Env             & 21.326   & 29.228    & 26.953     &26.777          &24.682         &24.542       &28.287     \\
\hline
w/o ILE             &35.090    &38.843     & 41.407     &38.046        &35.489        &34.544        &38.116            \\
w/o background             &35.131    &39.835     & 40.496     &37.649        &35.562        &34.893         &31.263            \\
w/o pre-convolution             & 35.072   & 39.608   & 41.559      &38.567        &35.987        &35.110         &38.336          \\
 w/o  regularization          & \cc{1}35.983    & \cc{1}41.439   & \cc{1}42.167    &\cc{1}39.004          &\cc{1}36.429       &\cc{1}35.535         &\cc{1}38.540
 \\  
 Ours               & \cc{2}35.296    & \cc{2}39.868   & \cc{2}41.735        &\cc{2}38.840         &\cc{2}36.301         &\cc{2}35.346         &\cc{2}38.506         \\\hline
\end{tabularx}
\label{tab:quan_glossy_psnr}
\end{table*}

\begin{table*}[h]
\centering
\caption{Per-Scene quantitative comparison of SSMIs on Glossy Blender dataset.}
\small
\begin{tabularx}{\linewidth}{@{}l|CCCCCCC}
\hline
                   & Car  &Helmet & Coffee  &RedBall    & GoldBall & GoldBall & MetalBall  \\
                   & Playroom  &Bedroom & Tearoom  &Playroom    & Bedroom & Bedroom2 & Tearoom  \\
\hline
 NeILF             & 0.908    & 0.942     &0.966      &0.917         &0.917          &0.913         &0.947         \\
 NeILF$^*$         & 0.926    & 0.955     &0.969       &0.924          &0.924 
         &0.922         &0.950   \\
Ne-Env             &0.926    & 0.954     & 0.969     &0.918         &0.917          &0.915      &0.946      \\
\hline
w/o ILE        &\cc{2}0.970     &0.981       &\cc{2}0.982      &0.982       &0.981         &0.978          &0.981 \\  
w/o background              &\cc{2}0.970     &0.988      & 0.981      &0.980       &0.982         &\cc{2}0.981          &0.971            \\
w/o pre-convolution             &0.969    &\cc{2}0.985    &\cc{2}0.982       &0.985         &0.982         &0.980         &\cc{1}0.982           \\
 w/o  regularization          & \cc{1}0.973     & \cc{1}0.990    & \cc{1}0.983    &\cc{1}0.986          &\cc{1}0.984      &\cc{1}0.982          &\cc{1}0.982 
 \\  
 Ours               &0.968     & \cc{2}0.985   &0.981    &\cc{1}0.986         &\cc{2}0.983          &\cc{2}0.981         &\cc{1}0.982         \\\hline
\end{tabularx}
\label{tab:quan_glossy_ssmi}
\end{table*}

\begin{table*}[!htbp]
\centering
\caption{Per-Scene quantitative comparison of LPIPS on Glossy Blender dataset.}
\small
\begin{tabularx}{\linewidth}{@{}l|CCCCCCC}
\hline
                   & Car  &Helmet & Coffee  &RedBall    & GoldBall & GoldBall & MetalBall  \\
                   & Playroom  &Bedroom & Tearoom  &Playroom    & Bedroom & Bedroom2 & Tearoom  \\
\hline
 NeILF             & 0.115    &0.081     &0.056       &0.096          &0.099          &0.108         &0.057         \\
 NeILF$^*$         &0.109    &0.073     &0.062       &0.094          &0.095          &0.098         &0.060  \\
Ne-Env             &0.110    &0.074     &0.063     &0.095           &0.098         &0.102        &0.062      \\
\hline
w/o ILE             &0.049      &0.047      &\cc{1}0.045      &0.025       &0.025        &0.027          &\cc{1}0.030            \\
w/o background              & \cc{2}0.048     &\cc{2}0.027      &0.048      &0.026       &\cc{1}0.019        &\cc{2}0.021          &0.038            \\
w/o pre-convolution             &0.054    & 0.034    & 0.046       &0.019            &0.021         &0.023         &\cc{1}0.030           \\
 w/o  regularization          & \cc{1}0.047     & \cc{1}0.025   & \cc{1}0.045     &\cc{1}0.017          &\cc{1}0.019        &\cc{1}0.020          &\cc{1}0.030 
 \\  
 Ours               &0.057           &0.036        & 0.046      &\cc{2}0.018          &0.021          &\cc{2}0.021          &\cc{1}0.030          \\\hline
\end{tabularx}
\label{tab:quan_glossy_lpips}
\end{table*}

\subsection{Discussion about the geometric perturbation.}

Previous studies have shown that the decomposition between light and material relies on the geometric quality \cite{park2020seeing, yao2022neilf, zhang2022modeling}.
In Section 4 of the main text, we adopt the rendered depth and normal maps as the result of stage one to isolate the influence of geometry quality. This means that the quality of novel-view synthesis will only reflect the lighting reconstruction and material decomposition.
However, in practice, there is always some deviation in geometry. Few studies have discussed the impact and underlying reason for this dependency. Through the experiment, we find that the decomposition of the mirror-like surfaces is more sensitive to the geometric perturbation, as shown in Fig. \ref{fig:geometry_disc}. This because our physically based lobe tracing relies on the reconstructed geometry to calculate the incoming direction. If the estimated geometry deviates from the true geometry, different tracing lobes may conflict with each other, which becomes more serious when the lobe is thinner. 
Despite this challenge, our method still produces better results of novel-view synthesis than existing methods, which demonstrates its robustness to geometric perturbation. 
Moreover, we suggest that this error could be potentially exploited to reconstruct a more accurate geometry in the future work.

\begin{figure*}[thb]
\centering
\includegraphics[width=0.8\linewidth]{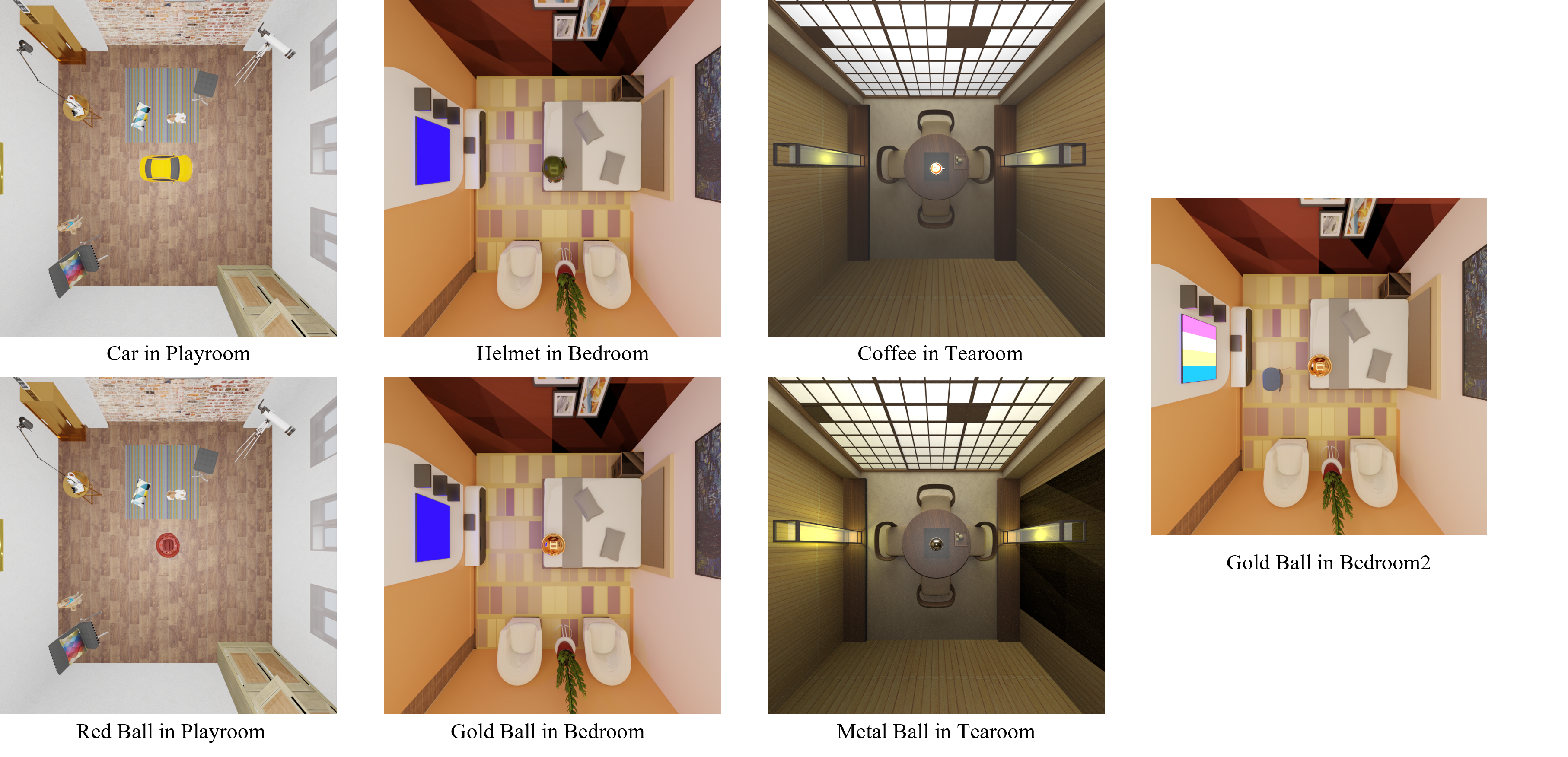}
\caption{Visualizations of the our new “Glossy Blender" scenes. Each one is rendered by a camera under the ceiling with a focal length of around 12mm.
 }
\label{fig:dataset}
\end{figure*}

\subsection{Dataset License}
We overview each scene in Fig. \ref{fig:dataset}. Our new ``Glossy Blender'' scenes were adapted from the following Sketchfab \href{https://sketchfab.com/feed}{https://sketchfab.com/feed} models, including object models and ambient illumination models. \\The object models are:
\begin{enumerate}
    \item Car: created by \emph{DAVID.3D.ART}, CC Attribution-NonCommercial license (model \emph{AC-Bmw 1m});
    \item Helmet: released by public NeILF dataset \cite{yao2022neilf};
    \item Coffee: created by \emph{Elen}, CC Attribution-ShareAlike license (model \emph{Coffee cup with plate});
    \item GoldBall \& RedBall \& MetaBall: released by public NeRF blender dataset \cite{mildenhall2020nerf} (model \emph{materials dataset}).
\end{enumerate}
The ambient illumination models are:
\begin{enumerate}
    \item Tearoom: created by \emph{Anex}, CC Attribution license (model \emph{Room});
    \item Playroom: created by \emph{enteRenders}, CC Attribution license (model \emph{Interior Living Room Corner});
    \item Bedroom \& Bedroom2: created by \emph{ankitk2618}, CC Attribution license (model \emph{Bedroom Interior}).
\end{enumerate}
We have slightly remixed, transformed, and built upon the material of models for better visual harmonization between objects and ambient illuminations.

\noindent\textbf{Acknowledgements. }
This work was supported by the NSFC grant 62001213, 62025108, and a gift funding from Tencent Rhino-Bird Research Program.

{\small
\bibliographystyle{ieee_fullname}
\bibliography{egbib}
}

\end{document}